\documentclass[runningheads]{llncs}

\usepackage{eccv}

\usepackage{eccvabbrv}

\usepackage{graphicx}
\usepackage{booktabs}
\usepackage{xcolor}
\usepackage{amsmath}
\usepackage{multirow}

\usepackage[accsupp]{axessibility}  

\usepackage{hyperref}

\usepackage{orcidlink}

\usepackage{tabularx}

\usepackage{pdfpages} 
\usepackage{pgffor} 

\makeatletter
\AtBeginDocument{\let\LS@rot\@undefined}
\makeatother

\def\supplementfilename{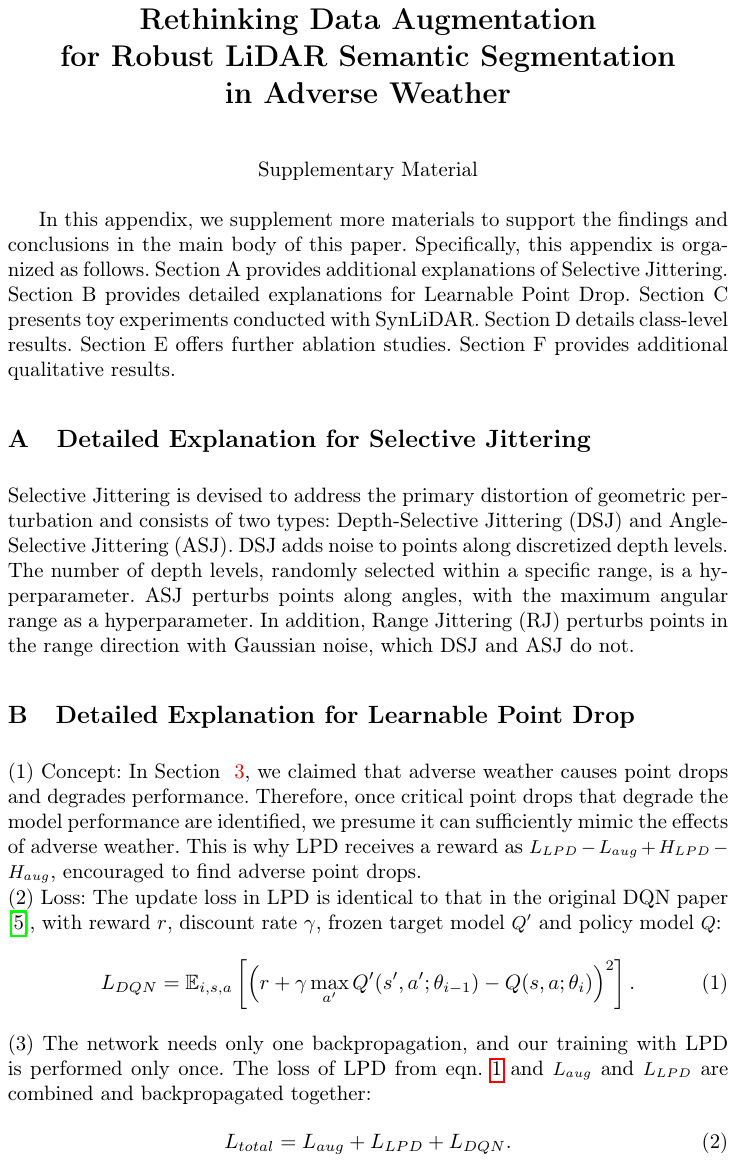}

\pdfximage{\supplementfilename}
\def\numbersupplementpages{\the\pdflastximagepages}

\newif\ifarXiv
\arXivtrue 

\begin{document}

\title{Rethinking Data Augmentation \\ for Robust LiDAR Semantic Segmentation \\ in Adverse Weather}

\titlerunning{Rethinking Data Augmentation for Robust LiDAR Semantic Segmentation}

\author{Junsung Park\textsuperscript{1} \and
Kyungmin Kim\textsuperscript{2} \and
Hyunjung Shim\textsuperscript{1}\thanks{Corresponding author: Hyunjung Shim}}


\authorrunning{J.Park et al.}

\institute{Korea Advanced Institute of Science \& Technology (KAIST) \and
Yonsei University \\
\email{jshackist@kaist.ac.kr, kyungmin.kim@yonsei.ac.kr, kateshim@kaist.ac.kr}\\
}

\maketitle

\begin{abstract}
    Existing LiDAR semantic segmentation methods often struggle with performance declines in adverse weather conditions. 
    Previous work has addressed this issue by simulating adverse weather or employing universal data augmentation during training.
    However, these methods lack a detailed analysis and understanding of how adverse weather negatively affects LiDAR semantic segmentation performance.
    Motivated by this issue, we identified key factors of adverse weather and conducted a toy experiment to pinpoint the main causes of performance degradation: (1) Geometric perturbation due to refraction caused by fog or droplets in the air and (2) Point drop due to energy absorption and occlusions.
    Based on these findings, we propose new strategic data augmentation techniques. First, we introduced a Selective Jittering (SJ) that jitters points in the random range of depth (or angle) to mimic geometric perturbation. 
    Additionally, we developed a Learnable Point Drop (LPD) to learn vulnerable erase patterns with a Deep Q-Learning Network to approximate the point drop phenomenon from adverse weather conditions.
    Without precise weather simulation, these techniques strengthen the LiDAR semantic segmentation model by exposing it to vulnerable conditions identified by our data-centric analysis. 
    Experimental results confirmed the suitability of the proposed data augmentation methods for enhancing robustness against adverse weather conditions. 
    Our method achieves a notable 39.5 mIoU on the SemanticKITTI-to-SemanticSTF benchmark, improving the baseline by 8.1\%p and establishing a new state-of-the-art. 
    Our code will be released at \url{https://github.com/engineerJPark/LiDARWeather}.
  \keywords{LiDAR Semantic Segmentation \and Robust Learning for Adverse Weather \and Data Augmentation}
\end{abstract}


\section{Introduction}
\label{sec:introduction}

LiDAR semantic segmentation is a fundamental task of 3D scene understanding, particularly in safety-critical applications like autonomous driving. 
However, existing LiDAR segmentation models~\cite{choy_4d_2019,ando_rangevit_2023,lai_spherical_2023,thomas_kpconv_2019,puy_using_2023,zhu_cylindrical_2021} commonly lack robustness, showing significant performance degradation under adverse weather conditions such as snow, fog, rain, or wet surfaces.

To address this issue, recent studies~\cite{kong_robo3d_2023, xiao_3d_2023, yan_benchmarking_2024} have introduced corruption benchmarks for adverse weather and proposed effective techniques for robust LiDAR segmentation against corruption. These robust methods are primarily divided into task-agnostic and simulation-based approaches.
The task-agnostic approach~\cite{xiao_3d_2023,kong_robo3d_2023,saltori2022cosmix} employs general machine-learning techniques for robustness without explicitly addressing LiDAR corruption caused by adverse weather.
The simulation-based approach~\cite{hahner_fog_2021, bijelic_benchmark_2018, qian_robust_2021, hahner_lidar_2022, kilic_lidar_2021, yang_realistic_2023, shin_characteristics_2019} artificially synthesizes weather-specific data using physical equations for training. 
However, these efforts focus on detection tasks and consider only a single type of weather at a time.

While the simulation-based approach leverages the intrinsic properties of LiDAR scans under target adverse weather conditions, simulating every weather type at all possible severities is impractical and often inaccurate. 
Instead, we adopt a data-centric perspective to analyze corrupted LiDAR data. 
For example, distortions caused by rain, snow, and fog often create similar patterns, as simulated in \cite{kilic_lidar_2021}. 
That is, all those adverse weather shows point drop patterns due to attenuation or occlusion by droplet \cite{kilic_lidar_2021, hahner_lidar_2022, fersch_influence_2016}. 
While the simulation-based approach requires explicit modeling of the LiDAR for each weather condition, our data-centric approach potentially addresses complex distortions with a few unified patterns. 
Through existing research and our comprehensive analysis, we find that various adverse weather conditions create similar distortion patterns in LiDAR data. 
Based on our analysis, we categorize these distortions into two types: (1) geometric perturbation and (2) point drop. 
Our toy experiments demonstrate that these distortions are closely related to segmentation performance degradation.

Rooted by the toy experiment, we introduce two novel and strategic data augmentation methods tailored to the LiDAR distortion caused by adverse weather. 
By incorporating these augmentations during training, we aim to bolster the model's robustness for each distortion type. 
For geometric perturbation, we introduce Selective Jittering (SJ). This method applies Additive Gaussian Noise (AGN) to alter the XYZ-coordinate and intensity within a selective local region. 
To handle a point drop pattern, we developed a Learnable Point Drop (LPD) that employs a Deep Q-Network (DQN) to strategically remove points.
Our augmentation strategies are inspired by existing studies~\cite{kilic_lidar_2021, li_what_2021, smith_modeling_2018}, which shows geometric perturbation typically involves small, random alterations to the original points. 
This insight led to our choice of jittering as a suitable technique for addressing the geometric perturbation. 
Also, our augmentation strategy for point drop is motivated by existing studies~\cite{hahner_fog_2021, kilic_lidar_2021, hahner_lidar_2022} which state that intensity and depth of LiDAR beams play a significant role. This prompts us to use a DQN to detect point drops and adapt to such point drop patterns.

Finally, we apply the proposed augmentations when training the LiDAR segmentation model with a clean source dataset. Experimental results demonstrate that our method achieves new state-of-the-art performance on SemanticSTF, a real adverse weather dataset. 
Specifically, we achieved an impressive +8.1\%p mIoU gain over the baseline MinkowskiNet, nearly tripling the +2.5\%p mIoU improvement from previous work~\cite{xiao_3d_2023}. 
Our method achieved remarkable performance gains, ranging from +0.4 to +10.3\%p mIoU, across various model architectures (Minkowski, CENet, SPVCNN) and corruption dataset (SemanticSTF and SemanticKITTI-C). These results demonstrate the generalizability of the proposed augmentation methods. 


In summary, this paper presents several key contributions.
\begin{itemize}
    \item We identify two prevalent distortion types in LiDAR data caused by adverse weather, leading to performance degradation, through a data-centric analysis.
    \item We introduce two novel data augmentations tailored to each identified distortion type.
    \item Our method sets new state-of-the-art benchmarks on the SemanticKITTI to SemanticSTF benchmark. Notably, our improvement was +8.1\%p mIoU without relying on precise simulations of adverse weather in LiDAR point inputs. This represents a tripling of the improvement over the baseline, compared to what previous methods have achieved \cite{xiao_3d_2023}.
\end{itemize}

\section{Related Works}

\subsection{LiDAR Semantic Segmentation}
Existing 3D LiDAR point cloud semantic segmentation methods can be categorized into three types based on the data representation: Point-based, Projection-based, and Voxel-based.

Point-based methods~\cite{qi_pointnet_2017,thomas_kpconv_2019,zhao_point_2020} utilize the 3D points directly as input. 
KPConv \cite{thomas_kpconv_2019} initially clustering local points, aggregating these local features, and then feeding them into kernel point convolutions.
The Point Transformer\cite{zhao_point_2020} utilizes a transformer architecture to compute query points in each local region, obtained through k-nearest neighbors (kNN). 
The Point-Mixer \cite{avidan_pointmixer_2022} tried to adapt the MLP-Mixer \cite{tolstikhin_mlp-mixer_2021} for point cloud applications.
They achieve high performance but suffer from high computational costs due to the utilization of large-scale raw LiDAR data.

Projection-based methods~\cite{milioto2019rangenet++,ando_rangevit_2023,kong_rethinking_2023} project LiDAR points into a 2D image and performs the semantic segmentation using architectures successful in a 2D image.
RangeViT \cite{ando_rangevit_2023} directly adopts a ViT model pre-trained on 2D images, demonstrating that pre-trained power in 2D images can be effective as prior knowledge in range images. 
RangeFormer \cite{kong_rethinking_2023} proposes ``RangeAug'' to maximize the utility of range images created by projecting into 2D, producing multiple range image data to overcome the low performance in range image models.
Projection-based methods achieve fast inference speed but present a sub-optimal performance due to missing information after the projection.

Voxel-based methods~\cite{choy_4d_2019,zhu_cylindrical_2021,lai_spherical_2023} perform efficient computations by dividing 3D space into a voxel grid and aggregating points within the same voxel.
MinkowskiNet~\cite{choy_4d_2019} voxelizes LiDAR points with a cubic grid and applies sparse convolution. 
Cylinder3D~\cite{zhu_cylindrical_2021} proposes cylindrical partitions, reflecting LiDAR's characteristic that the density of points depends on the distance.
SphereFormer~\cite{lai_spherical_2023} utilizes a radial window and transformer architecture to aggregate long-range information and improve performance. 
Voxel-based methods achieve a balance between reasonable inference time and commendable segmentation performance.

\subsection{LiDAR Data Augmentation}
\label{sec:lidaraug}
Inspired by 2D image augmentation, conventional LiDAR segmentation methods apply classic scaling, rotation, flipping, and translation to augment LiDAR data. 
Recently, several out-of-context augmentation techniques~\cite{nekrasov_mix3d_2021,xiao2022polarmix,kong2023lasermix} that mix different LiDAR scans have been proposed. 
Mix3D~\cite{nekrasov_mix3d_2021} combines randomly selected two scans. 
Considering the sweeping mechanism of the LiDAR sensor, PolarMix~\cite{xiao2022polarmix} cuts LiDAR scans along the azimuth axis, then exchanges point cloud sectors and applies instance-level rotate-pasting.
To reflect the spatial prior of the LiDAR point cloud, LaserMix~\cite{kong2023lasermix} partitions LiDAR scans based on the laser beams and blends partitions from different LiDAR scans.
Recent study~\cite{ryu2023instant} introduces a fast LiDAR domain augmentation module to address sensor-bias problems.
To the best of our knowledge, our approach is the first augmentation method specifically designed to address data corruption under adverse weather conditions.

\subsection{LiDAR under Adverse Weather Conditions}
\label{sec:distortions}
Robustness under harsh conditions is crucial in safety-critical applications. Adverse weather substantially degrade performance in real-world outdoor autonomous driving. Thus, there are several attempts to develop weather-robust models in fields such as 2D segmentation~\cite{lee2022fifo,li2022weather}, 3D detection~\cite{hahner_fog_2021,hahner_lidar_2022,kong_robo3d_2023}, and 3D segmentation~\cite{xiao_3d_2023,kong_robo3d_2023}. 
Simulation-based approaches~\cite{hahner_fog_2021,hahner_lidar_2022} artificially synthesize data for single weather conditions through physical modeling and utilize it for training. We differ from these in that we do not model specific weather conditions explicitly.
Recently proposed task-agnostic approaches~\cite{xiao_3d_2023,kong_robo3d_2023,saltori2022cosmix} consider multiple weather conditions at once. However, they use general machine-learning techniques (such as teacher-student framework and feature prototype) to achieve robustness rather than specifically tackle LiDAR corruption caused by adverse weather. We differ from these in that we propose augmentation methods specifically tailored for adverse weather conditions based on the analysis of performance degradation in LiDAR data.

\section{Finding Distortions to Augment}
\label{sec:recent_adverse_weather}

In this section, we aim to discuss the patterns of distortion that different adverse weather conditions impose on LiDAR data. 
Although adverse weather conditions are distinct in reality, studies have shown that their effects on LiDAR data often result in similar impacts. 
For instance, the distortions caused by "rain", "snow", and "fog" tend to produce similar point-missing patterns due to attenuation in the data, as demonstrated in \cite{kilic_lidar_2021}. 
Therefore, this section will focus on identifying the common distortion patterns caused by adverse weather through existing studies.
Overall, existing studies describe the effect of adverse weather as four different types of distortion: (1) Point Drop due to energy absorption, (2) Occlusions caused by droplets of rain or snow and fog, (3) Geometric perturbation, and (4) Intensity distortion due to energy absorption.

\subsection{Distortion Factors from Adverse Weather}
\label{sec:31}

\noindent
\textbf{(D1) Point Drop.}
Several studies have explored how adverse weather conditions contribute to point drops in LiDAR data. 
Kilic \etal. \cite{kilic_lidar_2021}, Fersch \etal. \cite{fersch_influence_2016} and Shin \etal. \cite{shin_characteristics_2019} describe point drops resulting from beam attenuation and beam missing due to droplets, fog, and frozen or wet ground. 
These studies collectively suggest that adverse weather conditions typically lead to point drops in LiDAR data.

\noindent
\textbf{(D2) Occlusions.}
Several studies have addressed occlusions caused by adverse weather conditions. 
Hahner \etal. \cite{hahner_lidar_2022}, Kilic \etal. \cite{kilic_lidar_2021}, Kong \etal. \cite{kong_robo3d_2023} and Yan \etal. \cite{yan_benchmarking_2024} consider scenarios where beams colliding with snow collect signals at much shorter distances than the intended objects of collision. 
Upon reviewing these studies, we have concluded that adverse weather consistently leads to occlusions. 

\noindent
\textbf{(D3) Geometric Perturbation.}
Some studies focus on geometric perturbation caused by adverse weather conditions. 
Kilic \etal. \cite{kilic_lidar_2021}, Li \etal. \cite{li_what_2021} and Smith \etal. \cite{smith_modeling_2018} demonstrated geometric perturbation in adverse weather, such as fog, snow, and rain, by incorporating random noise into the coordinates.
Through these studies, we have come to conclude that adverse weather universally causes geometric perturbation.

\noindent
\textbf{(D4) Intensity Distortion.}
Numerous studies have focused on intensity distortion caused by adverse weather conditions. 
Bijelic \etal. \cite{bijelic_benchmark_2018}, Shin \etal. \cite{shin_characteristics_2019}, Fersch \etal. \cite{fersch_influence_2016}, Kong \etal. \cite{kong_robo3d_2023}, and Yan \etal. \cite{yan_benchmarking_2024} have collectively shown that adverse weather conditions like fog, wetness, and rain lead to a reduction in LiDAR beam intensity, influencing the generation of synthetic data.
Through these studies, we conclude that adverse weather commonly causes intensity distortion.

\subsection{Toy Experiment}
\label{sec:distortion_test}

\noindent
\setlength{\tabcolsep}{0.7mm}{
\begin{table}[t]
\centering
\caption{Results of toy experiments from \textit{validation set} of SemanticKITTI~\cite{behley_semantickitti_2019}. Soft and Hard indicate the severity of the distortions.}
\begin{footnotesize}
\scalebox{0.7}{
\begin{tabular}{l|ccccccccccccccccccc|c}
    \toprule
    Distortion & \rotatebox{90}{car} & \rotatebox{90}{bi.cle} & \rotatebox{90}{mt.cle} & \rotatebox{90}{truck} & \rotatebox{90}{oth-v.} & \rotatebox{90}{pers.} & \rotatebox{90}{bi.clst} & \rotatebox{90}{mt.clst} & \rotatebox{90}{road} & \rotatebox{90}{parki.} & \rotatebox{90}{sidew.} & \rotatebox{90}{oth-g.} & \rotatebox{90}{build.} & \rotatebox{90}{fence} & \rotatebox{90}{veget.} & \rotatebox{90}{trunk} & \rotatebox{90}{terra.} & \rotatebox{90}{pole} & \rotatebox{90}{traf.} & mIoU \\
    \midrule
    Clean & 96.8 & 22.2 & 66.9 & 89.1 & 65.2 & 67.0 & 84.7 & 0.0 & 93.8 & 50.6 & 81.4 & 0.1 & 91.1 & 63.0 & 88.1 & 67.7 & 74.5 & 63.8 & 47.7 & 63.9 \\
    \midrule
    \textbf{D1} : Soft & 96.3 & 17.9 & 63.4 & 87.1 & 63.7 & 64.3 & 81.4 & 0.0 & 92.7 & 47.5 & 79.7 & 0.1 & 90.1 & 62.3 & 87.3 & 64.2 & 74.1 & 61.1 & 41.6 & 61.8 \\
    \textbf{D1} : Hard & 86.1 & 0.2 & 12.7 & 21.0 & 40.0 & 10.1 & 24.4 & 0.0 & 4.2 & 10.2 & 11.1 & 0.0 & 71.7 & 47.8 & 77.0 & 34.6 & 27.6 & 30.4 & 17.6 & \textbf{27.7} \\
    \textbf{D2} : Soft & 94.8 & 16.3 & 51.7 & 66.3 & 53.9 & 59.2 & 52.1 & 0.0 & 92.5 & 44.9 & 79.2 & 0.1 & 89.7 & 61.4 & 87.3 & 63.2 & 74.3 & 59.2 & 40.2 & 57.2 \\
    \textbf{D2} : Hard & 81.3 & 0.5 & 5.4 & 13.1 & 35.5 & 7.6 & 0.7 & 0.0 & 2.6 & 8.5 & 10.8 & 0.0 & 71.1 & 45.7 & 76.3 & 32.9 & 27.3 & 26.7 & 16.1 & \textbf{24.3} \\
    \textbf{D3} : Soft & 96.2 & 15.4 & 56.7 & 58.6 & 51.3 & 51.8 & 78.1 & 0.0 & 66.3 & 33.9 & 46.1 & 0.0 & 86.1 & 61.6 & 84.4 & 62.0 & 56.8 & 61.5 & 44.0 & 53.2 \\
    \textbf{D3} : Hard & 93.8 & 9.7 & 38.3 & 19.4 & 31.4 & 35.2 & 55.2 & 0.0 & 9.9 & 12.3 & 12.2 & 0.0 & 48.9 & 40.8 & 71.4 & 55.5 & 31.2 & 58.7 & 40.8 & \textbf{35.0} \\
    \textbf{D4} : Soft & 96.3 & 21.7 & 61.1 & 89.4 & 61.8 & 68.9 & 83.2 & 0.0 & 92.9 & 44.4 & 79.1 & 0.1 & 90.4 & 56.7 & 86.9 & 68.2 & 71.1 & 63.9 & 48.3 & 62.3 \\
    \textbf{D4} : Hard & 95.0 & 17.2 & 50.4 & 81.4 & 57.0 & 64.9 & 79.7 & 0.0 & 90.7 & 38.8 & 64.7 & 0.9 & 88.4 & 45.3 & 83.5 & 62.8 & 50.5 & 60.1 & 47.9 & 56.8 \\
    \bottomrule
\end{tabular}
}
\label{tab:validation_test}
\end{footnotesize}
\end{table}


Based on previous research findings mentioned in Section \ref{sec:31}, the types of distortions in LiDAR point clouds caused by adverse weather conditions converge into a set of common distortions. 
Therefore, from a data-centric perspective, the issues that we need to consider can be summarized as follows:
\textbf{(D1) Point Drop},
\textbf{(D2) Occlusions},
\textbf{(D3) Geometric Perturbation}, and
\textbf{(D4) Intensity Distortion}.
Here, we aim to identify which distortion types negatively impact performance.
To achieve this, we have generated four distortion types of toy synthetic data through the SemanticKITTI \textit{validation set}. 

\begin{itemize}
    \item \textbf{(D1) Point Drop:}  
    It considers the scenario where each LiDAR point disappears randomly and independently due to severe adverse weather.
    We removed individual points randomly to synthesize these data.
    We set the drop ratios at 0.5 and 0.9. 
    
    \item \textbf{(D2) Occlusions:}
    The method assumes that occlusions occur predominantly in front of objects, mainly due to distortion from fog, snow, and rain.
    We synthesized this data by randomly selecting points and altering their depth to one-tenth of the original depth.
    The selecting ratios were also determined to be 0.5 and 0.9.

    \item \textbf{(D3) Geometric Perturbation:} 
    This aspect assumes the distortion of point coordinates resulting from adverse weather conditions.
    We synthesized data by adding Gaussian noise to the coordinates of all points.
    The Gaussian noise levels were established at 0.05 and 0.25.

    \item \textbf{(D4) Intensity Distortion:} 
    This distortion type assumes intensity attenuation caused by fog, raindrops, and snow particles.
    We synthesized this data by subtracting Gaussian noise from the intensity of all points.
    The Gaussian noise levels were established at 0.05 and 0.25.
\end{itemize}

We selected MinkowskiNet \cite{choy_4d_2019} for our toy experiment. 
This choice was based on existing research \cite{kong_robo3d_2023, yan_benchmarking_2024}, which indicates that MinkowskiNet is a standard and robust model. 

In Tab. \ref{tab:validation_test}, it's evident that with increasing distortion in D1, D2, and D3, performance falls below half of the baseline. 
Contrarily, D4 maintains performance levels despite significant distortion. 
This performance degradation likely stems from changes in the local geometric structure where model computations occur. 
For this reason, D4 did not significantly impact performance. 

Importantly, D2's notable performance drop is argued to result from a point drop akin to D1.
It is due to occlusions altering the point's local geometric structure, an effect similar to point drop. 
This is evidenced in Fig. \ref{fig:validation_test}, where incorrect predictions from D2 show patterns resembling those in D1.

In summary, our observations indicate that geometric perturbation and point drop are the most impactful distortions from adverse weather. 
Data augmentation that replicates these phenomena could likely boost the model's robustness against such conditions without explicit weather simulations. 
Thus, robust LiDAR semantic segmentation models against adverse weather involve two key steps: 
\textbf{(1) Training on specific adverse point drops detrimental to performance}, and 
\textbf{(2) Training with minor adjustments in point coordinates}. 

\begin{figure}[t]
\begin{center}
   \includegraphics[width=1.0\linewidth]{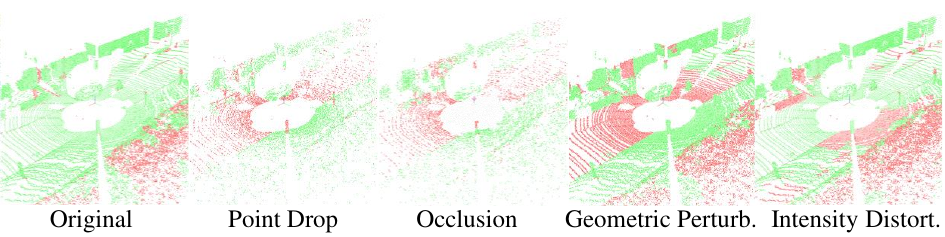}
\end{center}
   \caption{Visualization for the results of toy experiment. In all figures, green (or red) points indicate correct (or incorrect) predictions. Point drop, occlusion, and geometric perturbation are the main distortions that lead to performance degradation. Also, since the misprediction areas of point drop and occlusion largely overlap, it's reasonable to consider them as the same type of distortion.}
\label{fig:validation_test}
\end{figure}
\section{Methods}

In this section, we propose solutions for the two challenges identified in Section \ref{sec:distortion_test}.
Our objective is to enhance the robustness of the model by addressing the common distortions exhibited by various adverse weather conditions, through data augmentation. 
This approach aims to achieve robustness without the necessity of simulating each adverse weather scenario explicitly.

As observed earlier, distortions resulting from adverse weather include point drop, occlusion, geometric perturbation, and intensity distortion.
Of these, the main distortions identified as having the most detrimental impact on performance were (1) geometric perturbation caused by minor variations in point coordinates and (2) point drop caused by beam missing or occlusion.
To address these challenges, we propose two techniques: Selective Jittering and Learnable Point Drop.


\subsection{Selective Jittering}
Selective Jittering is devised to address the first main distortion, geometric perturbation.
Selective Jittering consists of two types: Depth-Selective Jittering (DSJ) and Angle-Selective Jittering (ASJ).

\noindent\textbf{Depth-Selective Jittering (DSJ).}
DSJ employs a single scan to add Gaussian noise to the XYZ coordinates and the intensity value of points below a randomly selected depth range. 

\noindent\textbf{Angle-Selective Jittering (ASJ).}
It applies Gaussian noise to a randomly selected range of angles.
By performing jittering across different ranges of depths or angles for each point, this method reflects the augmentation that some beams are affected by geometric perturbation in a non-uniform manner in adverse weather. 
This concept realistically represents the characteristics of transparent droplets like rain and snow \cite{fersch_influence_2016}. 

A significant aspect of DSJ and ASJ is that they use only a single frame. 
DSJ and ASJ are efficient in data augmentation without additional LiDAR frames and can perform reasonable augmentation even when the given LiDAR data is not sequentially captured. 

\noindent\textbf{Range Jittering (RJ).}
Range Jittering has been proposed to simulate range distortion caused by droplets and fog, as mentioned in \cite{kilic_lidar_2021}. Unlike DSJ and ASJ, Range Jittering applies jittering only in the range direction of points. This method is used in place of using the original points in DSJ and ASJ.

\subsection{Learnable Point Drop}

Learnable Point Drop (LPD) was devised to address the second main distortion, point drop.
LPD is designed to artificially create point drop due to occlusion, such as dense fog. 
LPD employs a Deep Q-Learning Network (DQN) \cite{mnih_human-level_2015} for identifying the drop ratio and drop region that lead to adverse effects on the model.
The reward of the DQN is designed to identify point drops that increase the training loss and uncertainty of the LiDAR semantic segmentation model. 
Through LPD, the LiDAR semantic segmentation model is exposed to point drop scenarios caused by adverse weather conditions. 
Consequently, it learns to make accurate predictions even in the absence of points that are critical for segmentation in clean data environments.
Since LPD is employed merely as a concept for data augmentation and exists as a separate module, it necessitates no alterations to the existing model's training scheme except for limiting the gradient norm to ensure the stability of DQN learning.

LPD module defines its current state by summing the loss $L_{aug}$, calculated from augmented data obtained through SJ, and the entropy $H_{aug}$, derived from logits. 
The loss $L_{aug}$ is calculated using the original loss function employed by the model in use.
The calculation of entropy is as follows:
{\small $$ H_{aug}(x) = -\frac{1}{N} \sum_{i=1}^{n} P(x_i) \log P(x_i) $$}
In this process, $x_i$ represents each point, and $P(\cdot)$ is softmax to the logits. $N$ denotes the number of points. 
LPD predicts the indices of points to drop, taking as input the sum of $L_{aug}$ and $H_{aug}$ added to the input point tensor.  
This process enables simultaneous learning of the drop ratio and drop region. 
Subsequently, the difference between the sum of the loss $L_{LPD}$ and entropy $H_{LPD}$, obtained from the dropped points by LPD, and the sum of $L_{aug}$ and $H_{aug}$, is defined as the reward.

\begin{figure}[t]
\begin{center}
   \includegraphics[width=1.0\linewidth]{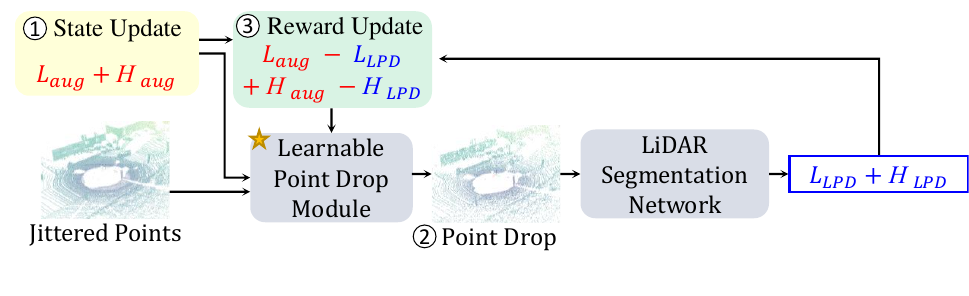}
\end{center}
   \caption{Description of Learnable Point Drop (LPD). (1) The LPD utilizes the loss $L_{aug}$ and entropy $H_{aug}$ derived from LiDAR segmentation model as its current state. (2) Upon receiving a point input, LPD performs a point drop, then recalculates the loss $L_{LPD}$ and entropy $H_{LPD}$. (3) The difference between these new values and the previous ones is used to compute the reward.}
\label{fig:LPD_proecss}
\end{figure}

In Comparison with random drop, random drop uniformly reduces points across all depths.
Consequently, this approach is not suitable for simulating point drops caused by fog, as it lacks the depth-specific characteristics of fog-related distortions.
Furthermore, randomly generated point drop scans may not achieve the severity of adverse weather conditions necessary to drop critical points that could deceive the model.
A detailed comparison will be conducted in the supplementary material.

\subsection{Overall Pipeline}

The overall training procedure begins with data augmentation through SJ. 
The augmented points are input into the LiDAR semantic segmentation model to compute loss. 
The model then calculates logits for each point, applies softmax, and subsequently computes the entropy. 
The current state of the LPD module is set as the sum of the loss and entropy. 
The dropped points are then input into the LiDAR semantic segmentation model to calculate the corresponding loss and entropy. 
The difference between the sum of loss/entropy obtained from the augmented points and that obtained from the points determined by LPD is set as the reward. 
Through this process, the LPD module enables the LiDAR semantic segmentation model to effectively learn points distorted due to adverse weather.
Overall process is detailed in Fig. \ref{fig:overall_process}

\begin{figure}
\begin{center}
   \includegraphics[width=1.0\linewidth]{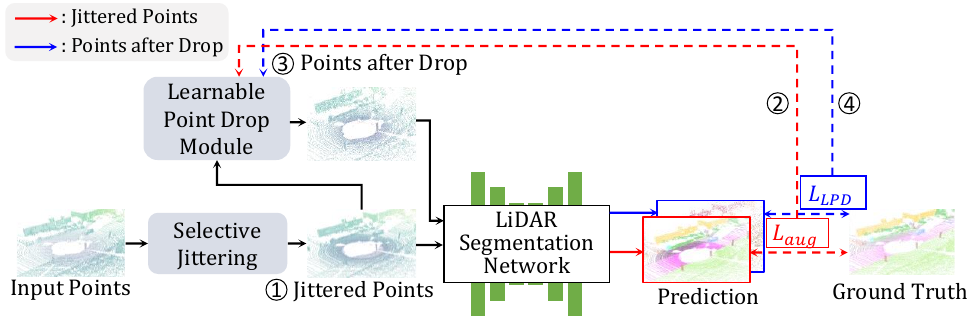}
\end{center}
   \caption{The overall training process. (1) Initially, SJ is applied. (2) Following this, we calculate the loss, which leads to the update of the LPD state. (3) This process is followed by utilizing LPD to generate dropped points. (4) Subsequent to recalculating the loss, the reward is accordingly updated.}
\label{fig:overall_process}
\end{figure}
\section{Experiments}

\subsection{Experimental Setup}
\noindent
\textbf{Dataset and Evaluation Metrics.}
To assess robustness, we trained our model on 19 classes from SemanticKITTI and then validated its performance on \textit{validation set} of SemanticSTF as same as \cite{xiao_3d_2023}. 
The experiment process involves training on clean data followed by evaluating robustness in adverse conditions, akin to existing research \cite{kong_robo3d_2023, xiao_3d_2023, yan_benchmarking_2024}.
SemanticSTF is a dataset gathered under real-world adverse conditions such as rain, fog, and snow. Following the approach in \cite{xiao_3d_2023}, invalid data in SemanticSTF are mapped to an ``ignore label''. 
We utilized MinkowskiNet \cite{choy_4d_2019} for validation purposes due to its well-known robustness \cite{kong_robo3d_2023, yan_benchmarking_2024}. 
Our evaluation metrics include the Intersection over Union (IoU) for each class and the mean IoU (mIoU) across all classes. 

\noindent
\textbf{Implementation Details.}
MinkowskiNet-18/32width served as our baseline model. The learning rate was set at 0.24, with a weight decay of 0.0001.
The mean and standard deviation of Gaussian noise used in SJ are 0 and 0.01 each. 
In cases of Learnable Point Drop, the norm of the gradient was limited to 100 for both the DQN and Segmentation model. 
All experiments were conducted using four A6000 GPUs for 15 epochs in batch size 2. The duration of all experiments ranged between 3 to 5 hours.

\subsection{Main Results}

\begin{table}[ht]
    \centering
    \caption{\footnotesize{LiDAR segmentation results (mIoU) on the SemanticSTF validation set of models trained with (a) SemanticKITTI and (b) SynLiDAR~\cite{xiao_transfer_2022}. D-fog and L-fog denote dense fog and light fog weather conditions in all experiments. $^\ast$ denotes the reproduced result with the same segmentation backbone. The symbol $\ddagger$ indicates that the validation for model selection was performed on sequence 0 of SynLiDAR, rather than SemanticSTF. Best score is in \textbf{bold} and second best is \underline{underlined}.}}
    \label{tab:ablation_all}
    \begin{minipage}[t]{0.49\linewidth}
        \centering
        \scriptsize
        \scalebox{0.85}{
        \begin{tabular}{l|cccc|c}
            \toprule
            \multicolumn{6}{c}{\scriptsize{(a) SemanticKITTI$\rightarrow$SemanticSTF}} \\
            \midrule
            \scriptsize{Methods} & \scriptsize{\rotatebox{90}{D-fog}} & \scriptsize{\rotatebox{90}{L-fog}} &  \scriptsize{\rotatebox{90}{Rain}} & \scriptsize{\rotatebox{90}{Snow}} & \scriptsize{mIoU} \\
            \midrule
            \scriptsize{Oracle} & \scriptsize{51.9} & \scriptsize{54.6} & \scriptsize{57.9} & \scriptsize{53.7} & \scriptsize{54.7}\\ 
            \midrule
            \scriptsize{Baseline} & \scriptsize{30.7} & \scriptsize{30.1} & \scriptsize{29.7} & \scriptsize{25.3} & \scriptsize{31.4} \\
            \scriptsize{LaserMix \cite{kong_lasermix_2023}} & \scriptsize{23.2} & \scriptsize{15.5} & \scriptsize{9.3} & \scriptsize{7.8} & \scriptsize{14.7} \\
            \scriptsize{PolarMix \cite{xiao_polarmix_nodate}} & \scriptsize{21.3} & \scriptsize{14.9} & \scriptsize{16.5} & \scriptsize{9.3} & \scriptsize{15.3} \\
            \scriptsize{PointDR$^\ast$ \cite{xiao_3d_2023}} & \scriptsize{\textbf{37.3}} & \scriptsize{\underline{33.5}} & \scriptsize{\underline{35.5}} & \scriptsize{\underline{26.9}} & \scriptsize{\underline{33.9}} \\
            \midrule
            \scriptsize{Baseline+SJ+LPD} & \scriptsize{\underline{36.0}} & \scriptsize{\textbf{37.5}} & \scriptsize{\textbf{37.6}} & \scriptsize{\textbf{33.1}} & \scriptsize{\textbf{39.5}} \\
            \scriptsize{{\textcolor{red}{Increments to baseline}}} & \scriptsize{{\textcolor{red}{+5.3}}} & \scriptsize{{\textcolor{red}{+7.4}}} & \scriptsize{{\textcolor{red}{+7.9}}} & \scriptsize{{\textcolor{red}{+7.8}}} & \scriptsize{{\textcolor{red}{+8.1}}} \\
            \bottomrule
        \end{tabular}
        }
    \end{minipage}
    \begin{minipage}[t]{0.49\linewidth}
        \centering
        \scriptsize
        \scalebox{0.85}{
        \begin{tabular}{l|cccc|c}
            \toprule
            \multicolumn{6}{c}{\scriptsize{(b) SynLiDAR$\rightarrow$SemanticSTF}} \\
            \midrule
            \scriptsize{Methods} & \scriptsize{\rotatebox{90}{D-fog}} & \scriptsize{\rotatebox{90}{L-fog}} &  \scriptsize{\rotatebox{90}{Rain}} & \scriptsize{\rotatebox{90}{Snow}} & \scriptsize{mIoU} \\
            \midrule
            \scriptsize{Oracle} & \scriptsize{51.9} & \scriptsize{54.6} & \scriptsize{57.9} & \scriptsize{53.7} & \scriptsize{54.7}\\ 
            \midrule
            \scriptsize{Baseline} & \scriptsize{15.24} & \scriptsize{15.97} & \scriptsize{16.83} & \scriptsize{12.76} & \scriptsize{15.45} \\
            \scriptsize{LaserMix \cite{kong_lasermix_2023}} & \scriptsize{15.32} & \scriptsize{17.95} & \scriptsize{18.55} & \scriptsize{13.8} & \scriptsize{16.85} \\
            \scriptsize{PolarMix \cite{xiao_polarmix_nodate}} & \scriptsize{16.47} & \scriptsize{18.69} & \scriptsize{19.63} & \scriptsize{15.98} & \scriptsize{18.09} \\
            \scriptsize{PointDR$^\ast$ \cite{xiao_3d_2023}} & \scriptsize{\underline{19.09}} & \scriptsize{20.28} & \scriptsize{\underline{25.29}} & \scriptsize{\underline{18.98}} & \scriptsize{19.78} \\
            \scriptsize{PointDR$^\ast$$\ddagger$} & \scriptsize{\textbf{21.41}} & \scriptsize{\underline{20.94}} & \scriptsize{\textbf{25.48}} & \scriptsize{\textbf{19.31}} & \scriptsize{\underline{20.47}} \\
            \midrule
            \scriptsize{Baseline+SJ+LPD} & \scriptsize{19.08} & \scriptsize{20.65} & \scriptsize{21.97} & \scriptsize{17.27} & \scriptsize{20.08} \\
            \scriptsize{{\textcolor{red}{Increments to baseline}}} & \scriptsize{{\textcolor{red}{+3.8}}} & \scriptsize{{\textcolor{red}{+4.7}}} & \scriptsize{{\textcolor{red}{+5.1}}} & \scriptsize{{\textcolor{red}{+4.5}}} & \scriptsize{{\textcolor{red}{+4.6}}} \\
            \scriptsize{Baseline+SJ+LPD$\ddagger$} & \scriptsize{18.99} & \scriptsize{\textbf{21.22}} & \scriptsize{23.14} & \scriptsize{17.28} & \scriptsize{\textbf{20.51}} \\
            \scriptsize{{\textcolor{red}{Increments to baseline}}} & \scriptsize{{\textcolor{red}{+3.7}}} & \scriptsize{{\textcolor{red}{+5.3}}} & \scriptsize{{\textcolor{red}{+6.3}}} & \scriptsize{{\textcolor{red}{+4.5}}} & \scriptsize{{\textcolor{red}{+5.1}}} \\
            \bottomrule
        \end{tabular}
        }
    \end{minipage}
    \label{tab:main_experiment}
\end{table}

\noindent
\textbf{SemanticKITTI to SemanticSTF.}
According to Tab. \ref{tab:main_experiment} (a), our proposed method showed a significant improvement in SemanticSTF, with a +8.1 mIoU increase over the baseline and +5.6 mIoU over the recent competitor, PointDR. 
Our method demonstrated universally superior performance across all weather conditions, validating the validity of our choice in data distortion type and the effectiveness of our method.
Our method substantially improved the performance of categories like \textit{other vehicle}, \textit{motorcyclist}, \textit{sidewalk}, \textit{pole}, and \textit{traffic sign}, which had lower baseline performance, increasing their class IoU by about +5 to +10. 
Additionally, it significantly enhanced the mIoU for cars and persons by +19.0 and +9.6, respectively, compared to the baseline, which is crucial for safety in driving environments. 
More details are available in the supplementary material.

Also, in Tab. \ref{tab:main_experiment} (a), examining the performance improvement per weather condition, we note that the most significant increase in mIoU was in rain, with +7.9 mIoU. 
This result is indicative of our SJ augmentation effectively reflecting geometric perturbations caused by raindrops. This aspect will be further explored in the ablation study.
Additionally, in snow, our method showed a remarkable increase of +7.8 mIoU, and in dense fog (+5.3 mIoU) and light fog (+7.4 mIoU), our methodology clearly outperformed the baseline. 
This indicates that our data augmentation effectively models the adverse impacts of weather on LiDAR semantic segmentation models. 
Detailed analysis of how our method enhances robustness against each weather condition, and the factors contributing to the improved performance, will be discussed in the ablation study.

Moreover, as previously mentioned, the training architectures designed to enhance robustness, such as those proposed in PointDR \cite{kong_robo3d_2023, yan_benchmarking_2024, xiao_3d_2023}, can be orthogonally applied to our proposed data augmentation. 
Therefore, regardless of the choice of training architecture, the integration of our method is expected to yield superior performance compared to conventional approaches.

\begin{figure}
\begin{center}
   \includegraphics[width=1.0\linewidth]{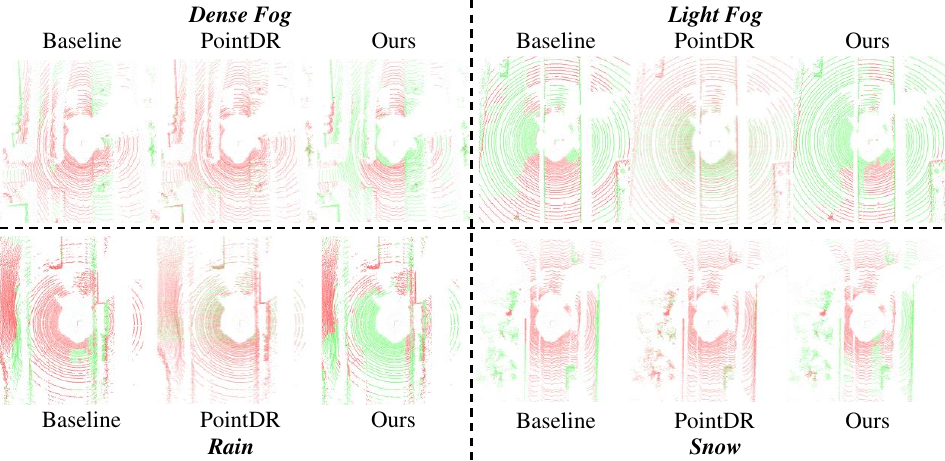}
\end{center}
   \caption{Qualitative results of our method from \textit{validation set} of SemanticSTF. All models are trained in \textit{train set} of SemanticKITTI. Green (or red) points indicate correct (or incorrect) predictions, showing our predictions are more accurate than PointDR in all weather, especially on \textit{road} and \textit{sidewalk}.}
\label{fig:stress_test}
\end{figure}

\noindent\textbf{SynLiDAR to SemanticSTF.}
For SynLiDAR, unlike SemanticKITTI, where no specific validation set was available, sequence 0 was treated as the validation set. 
The best-performing models during these validation checkpoints were selected for reporting. 

In contrast, PointDR demonstrated an unrealistic scenario in its baseline code, using SemanticSTF for validation during training and reporting only the highest-performing results.
For a fair comparison using PointDR's criteria, our method shows a performance increase of +0.04 mIoU over PointDR. Using our more realistic protocol, our method demonstrates an even greater performance improvement of +0.32 mIoU against PointDR.
Detailed results are shown in Tab. \ref{tab:main_experiment} (b).

\subsection{Additional Experiments}

\noindent
\textbf{Analysis of Proposed Component Methods.} 
As shown in As Tab. \ref{tab:componenet_analysis}, when ASJ was added to the baseline, we observed an improvement in robustness across all adverse weather conditions (increases of +2.6 mIoU in dense fog, +5.3 mIoU in light fog, +8.1 mIoU in rain, and +6.3 mIoU in snow), albeit with a -1.8 mIoU decrease in clean data performance. 
The introduction of DSJ further enhanced performance in dense fog (+4.1 mIoU), light fog (+6.3 mIoU), rain (+9.3 mIoU), snow (+4.6 mIoU), and an overall performance increase of +6.2 mIoU.
Our Selective Jittering method's strong performance in rain conditions showcases its efficacy in rainy weather, where it considers geometric perturbations for selected beams instead of all.

The application of Range Jittering showed improvements in dense fog(+3.7 mIoU), light fog (+6.9 mIoU), rain(+6.0 mIoU), snow (+8.2 mIoU), and an overall performance increase of +7.3 mIoU, indicating the importance of simulating range distortion due to adverse weather as discussed in \cite{kilic_lidar_2021}. 
As seen earlier, in rainy weather, jittering only some beams aids in enhancing robustness. 
However, performing Range Jittering on the remaining original points leads to all points being jittered, reducing performance in rain conditions. 
Additionally, the improvement in snow conditions by using Range Jittering suggests that perturbing as many beams as possible is more effective for robustness enhancement in snowy weather.

The addition of LPD resulted in improvements in dense fog (+5.3 mIoU), light fog (+7.4 mIoU), rain (+7.9 mIoU), snow (+7.8 mIoU), and an overall increase of +8.1 mIoU against to the baseline.
The substantial performance enhancements observed in conditions of dense and light fog indicate that the LPD effectively represents occlusion from foggy weather as intended.
The performance reduction in snow conditions compared to the use of ASJ, DSJ, and Range Jittering (-0.4 mIoU) is likely due to LPD not exclusively targeting the frozen or wet ground for point dropout.
LPD applies the point drop across all points, as shown in Fig. \ref{fig:LPD_result}. 
This represents a limitation of LPD, necessitating the development of alternative methods to address this issue.

Overall, the use of all components led to a significant improvement in performance under adverse weather conditions, with an increase of +8.1 mIoU over the baseline at a reasonable cost of -1.1 mIoU in clean data. 
Ultimately, our method achieved state-of-the-art results on SemanticSTF, with an increase in mIoU across all adverse weather conditions compared to the baseline. 
The combined use of DSJ, ASJ, Range Jittering, and LPD demonstrated high performance, indicating that each component synergistically contributes to enhancing the model's robustness.

\noindent
\setlength{\tabcolsep}{0.7mm}{
\begin{table}[t]
\centering
\caption{Experiments on the components of our methods with SemanticKITTI as source and SemanticSTF as target. The values in parentheses indicate the performance improvement or decrease over the baseline model.}
\begin{footnotesize}
\begin{tabular}{l|c|cccc|c}
    \toprule
    Methods & \rotatebox{90}{Clean} & \rotatebox{90}{D-fog} & \rotatebox{90}{L-fog} &  \rotatebox{90}{Rain} & \rotatebox{90}{Snow} & mIoU \\
    \midrule
    Baseline & \textbf{63.9} & 30.7 & 30.1 & 29.7 & 25.3 & 31.4 \\
    +ASJ & 62.1 {\tiny \textcolor{blue}{(-1.8)}} & 33.3 & 35.4 & \underline{37.8} & 31.6 & 36.8 {\tiny \textcolor{red}{(+5.4)}} \\
    +DSJ & \underline{63.0} {\tiny \textcolor{blue}{(-0.9)}} & \underline{34.8} & 36.4 & \textbf{39.0} & 29.9 & 37.6 {\tiny \textcolor{red}{(+6.2)}} \\
    +RJ & 61.2 {\tiny \textcolor{blue}{(-2.7)}} & 33.4 & \underline{37.0} & 35.7 & \textbf{33.5} & \underline{38.7} {\tiny \textcolor{red}{(+7.3)}} \\
    +LPD & 62.8 {\tiny \textcolor{blue}{(-1.1)}} & \textbf{36.0} & \textbf{37.5} & 37.6 & \underline{33.1} & \textbf{39.5} {\tiny \textcolor{red}{(+8.1)}} \\
    \bottomrule
\end{tabular}
\label{tab:componenet_analysis}
\end{footnotesize}
\end{table}

\begin{figure}
\begin{center}
   \includegraphics[width=0.6\linewidth]{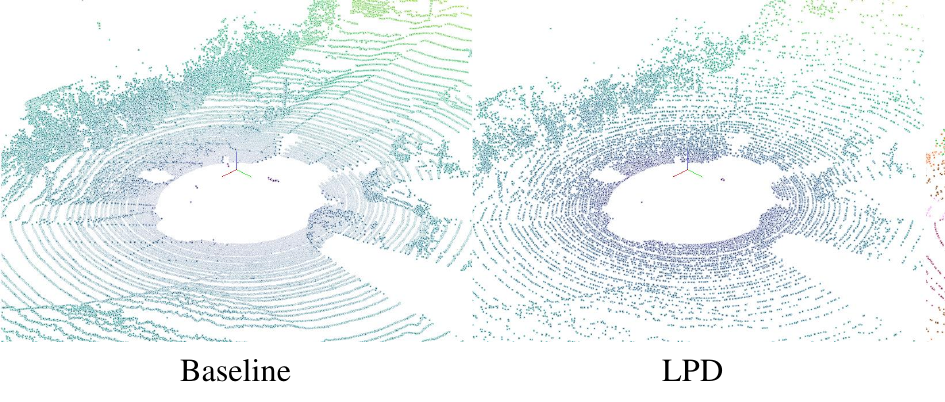}
\end{center}
   \caption{Qualitative results of LPD from \textit{validation set} of SemanticKITTI. It can be observed that the point drop facilitated by the LPD occurs not just locally but extends across all points. While LPD effectively mimics the occlusions and sparse point distributions seen in real adverse weather, it does not replicate the effect of wet ground surfaces.}
\label{fig:LPD_result}
\end{figure}

\noindent\textbf{Experiments on Various Input Data Representation, Models and Datasets.}
We evaluated our method using different data representations, architectures, and datasets, specifically: (1) projection-based CENet and (2) voxel-based SPVCNN for backbones, and (3) the synthetic corruption dataset SemanticKITTI-C \cite{kong_robo3d_2023}. Tab. 
~\ref{tab:cenet_spvcnn_semantickittic} shows our results.

SemanticKITTI-C has a unique data distribution with more samples in classes, such as \emph{sidewalks} and \emph{vegetation}, than SemanticSTF.
Our method consistently improves performance across various datasets and architectures.
Specifically, for the SemanticSTF dataset, our method increases the mIoU of CENet by +7.8, SPVCNN by +10.3, and MinkowskiNet by +8.1 mIoU. Similarly, for the SemanticKITTI-C dataset, the improvements are +3.9 for CENet, +0.4 for SPVCNN, and +5.6 mIoU for MinkowskiNet.
These consistent enhancements across different input data representations, models, and datasets highlight the robustness and generalizability of our approach.
\begin{table}[t]
    \centering
    \caption{LiDAR segmentation results (mIoU) with the various architecture and datasets. All models were trained with SemanticKITTI.}
    \begin{tabular}{ccc}
        \toprule
        Method & SemanticSTF & SemanticKITTI-C \\
        \midrule
        CENet & 14.2 & 49.3 \\
        SPVCNN & 28.1 & 52.5 \\
        Minkowski & 31.4 & 53.0 \\
        \midrule
        CENet+Ours & 22.0 { \textcolor{red}{(+7.8)}} & 53.2 { \textcolor{red}{(+3.9)}}\\
        SPVCNN+Ours & 38.4 { \textcolor{red}{(+10.3)}} & 52.9 { \textcolor{red}{(+0.4)}}\\
        Minkowski+Ours & 39.5 { \textcolor{red}{(+8.1)}} & 58.6 { \textcolor{red}{(+5.6)}}\\
        \bottomrule
    \end{tabular}
\label{tab:cenet_spvcnn_semantickittic}
\end{table}

\section{Conclusion}

Our study focused on analyzing LiDAR data distortions in adverse weather and enhancing LiDAR model robustness with a data-centric method. 
We identified that point drop and geometric distortion mainly affect model performance through toy experiments. 
We introduced novel data augmentation techniques, Selective Jittering, and Learnable Point Drop, leading to state-of-the-art performance on SemanticSTF.
We showed that our methodology significantly contributes to robustness enhancement without complex data simulations. 
We expect that our research will contribute to improving the stability and reliability of LiDAR semantic segmentation.

\noindent\textbf{Acknowledgements}. 
This work was supported by IITP grant funded by MSIT (No.2021-0-02068, Artificial Intelligence Innovation Hub and RS-2019-II190075, Artificial Intelligence Graduate School Program(KAIST)) and KEIT grant funded by MOTIE (No.2022-0-00680, No.2022-0-01045), NRF funded by the MSIP (NRF-2022R1A2C3011154, RS-2023-00219019, RS-2023-00240135) and Samsung Electronics Co., Ltd (IO230508-06190-01).


%
%
\bibliographystyle{splncs04}
\bibliography{main}

\begin{thebibliography}{10}
\providecommand{\url}[1]{\texttt{#1}}
\providecommand{\urlprefix}{URL }
\providecommand{\doi}[1]{https://doi.org/#1}

\bibitem{ando_rangevit_2023}
Ando, A., Gidaris, S., Bursuc, A., Puy, G., Boulch, A., Marlet, R.: Rangevit: Towards vision transformers for 3d semantic segmentation in autonomous driving. In: Proceedings of the IEEE/CVF conference on computer vision and pattern recognition. pp. 5240--5250 (2023)

\bibitem{behley_semantickitti_2019}
Behley, J., Garbade, M., Milioto, A., Quenzel, J., Behnke, S., Stachniss, C., Gall, J.: Semantickitti: A dataset for semantic scene understanding of lidar sequences. In: Proceedings of the IEEE/CVF international conference on computer vision. pp. 9297--9307 (2019)

\bibitem{bijelic_benchmark_2018}
Bijelic, M., Gruber, T., Ritter, W.: A benchmark for lidar sensors in fog: Is detection breaking down? In: 2018 IEEE intelligent vehicles symposium (IV). pp. 760--767. IEEE (2018)

\bibitem{avidan_pointmixer_2022}
Choe, J., Park, C., Rameau, F., Park, J., Kweon, I.S.: Pointmixer: Mlp-mixer for point cloud understanding. In: European Conference on Computer Vision. pp. 620--640. Springer (2022)

\bibitem{choy_4d_2019}
Choy, C., Gwak, J., Savarese, S.: 4d spatio-temporal convnets: Minkowski convolutional neural networks. In: Proceedings of the IEEE/CVF conference on computer vision and pattern recognition. pp. 3075--3084 (2019)

\bibitem{fersch_influence_2016}
Fersch, T., Buhmann, A., Koelpin, A., Weigel, R.: The influence of rain on small aperture lidar sensors. In: 2016 German Microwave Conference (GeMiC). pp. 84--87. IEEE (2016)

\bibitem{hahner_lidar_2022}
Hahner, M., Sakaridis, C., Bijelic, M., Heide, F., Yu, F., Dai, D., Van~Gool, L.: Lidar snowfall simulation for robust 3d object detection. In: Proceedings of the IEEE/CVF conference on computer vision and pattern recognition. pp. 16364--16374 (2022)

\bibitem{hahner_fog_2021}
Hahner, M., Sakaridis, C., Dai, D., Van~Gool, L.: Fog simulation on real lidar point clouds for 3d object detection in adverse weather. In: Proceedings of the IEEE/CVF international conference on computer vision. pp. 15283--15292 (2021)

\bibitem{kilic_lidar_2021}
Kilic, V., Hegde, D., Sindagi, V., Cooper, A.B., Foster, M.A., Patel, V.M.: Lidar light scattering augmentation (lisa): Physics-based simulation of adverse weather conditions for 3d object detection. arXiv preprint arXiv:2107.07004  (2021)

\bibitem{kong_rethinking_2023}
Kong, L., Liu, Y., Chen, R., Ma, Y., Zhu, X., Li, Y., Hou, Y., Qiao, Y., Liu, Z.: Rethinking range view representation for lidar segmentation. In: Proceedings of the IEEE/CVF International Conference on Computer Vision. pp. 228--240 (2023)

\bibitem{kong_robo3d_2023}
Kong, L., Liu, Y., Li, X., Chen, R., Zhang, W., Ren, J., Pan, L., Chen, K., Liu, Z.: Robo3d: Towards robust and reliable 3d perception against corruptions. In: Proceedings of the IEEE/CVF International Conference on Computer Vision. pp. 19994--20006 (2023)

\bibitem{kong2023lasermix}
Kong, L., Ren, J., Pan, L., Liu, Z.: Lasermix for semi-supervised lidar semantic segmentation. In: Proceedings of the IEEE/CVF Conference on Computer Vision and Pattern Recognition. pp. 21705--21715 (2023)

\bibitem{kong_lasermix_2023}
Kong, L., Ren, J., Pan, L., Liu, Z.: Lasermix for semi-supervised lidar semantic segmentation. In: Proceedings of the IEEE/CVF Conference on Computer Vision and Pattern Recognition. pp. 21705--21715 (2023)

\bibitem{lai_spherical_2023}
Lai, X., Chen, Y., Lu, F., Liu, J., Jia, J.: Spherical transformer for lidar-based 3d recognition. In: Proceedings of the IEEE/CVF Conference on Computer Vision and Pattern Recognition. pp. 17545--17555 (2023)

\bibitem{lee2022fifo}
Lee, S., Son, T., Kwak, S.: Fifo: Learning fog-invariant features for foggy scene segmentation. In: Proceedings of the IEEE/CVF Conference on Computer Vision and Pattern Recognition. pp. 18911--18921 (2022)

\bibitem{li_what_2021}
Li, Y., Duthon, P., Colomb, M., Ibanez-Guzman, J.: What happens for a tof lidar in fog? IEEE Transactions on Intelligent Transportation Systems  \textbf{22}(11),  6670--6681 (2020)

\bibitem{li2022weather}
Li, Z., Wu, X., Wang, J., Guo, Y.: Weather-degraded image semantic segmentation with multi-task knowledge distillation. Image and Vision Computing  \textbf{127},  104554 (2022)

\bibitem{milioto2019rangenet++}
Milioto, A., Vizzo, I., Behley, J., Stachniss, C.: Rangenet++: Fast and accurate lidar semantic segmentation. In: 2019 IEEE/RSJ international conference on intelligent robots and systems (IROS). pp. 4213--4220. IEEE (2019)

\bibitem{mnih_human-level_2015}
Mnih, V., Kavukcuoglu, K., Silver, D., Rusu, A.A., Veness, J., Bellemare, M.G., Graves, A., Riedmiller, M., Fidjeland, A.K., Ostrovski, G., et~al.: Human-level control through deep reinforcement learning. nature  \textbf{518}(7540),  529--533 (2015)

\bibitem{nekrasov_mix3d_2021}
Nekrasov, A., Schult, J., Litany, O., Leibe, B., Engelmann, F.: Mix3d: Out-of-context data augmentation for 3d scenes. In: 2021 international conference on 3d vision (3dv). pp. 116--125. IEEE (2021)

\bibitem{puy_using_2023}
Puy, G., Boulch, A., Marlet, R.: Using a waffle iron for automotive point cloud semantic segmentation. In: Proceedings of the IEEE/CVF International Conference on Computer Vision. pp. 3379--3389 (2023)

\bibitem{qi_pointnet_2017}
Qi, C.R., Su, H., Mo, K., Guibas, L.J.: Pointnet: Deep learning on point sets for 3d classification and segmentation. In: Proceedings of the IEEE conference on computer vision and pattern recognition. pp. 652--660 (2017)

\bibitem{qian_robust_2021}
Qian, K., Zhu, S., Zhang, X., Li, L.E.: Robust multimodal vehicle detection in foggy weather using complementary lidar and radar signals. In: Proceedings of the IEEE/CVF Conference on Computer Vision and Pattern Recognition. pp. 444--453 (2021)

\bibitem{ryu2023instant}
Ryu, K., Hwang, S., Park, J.: Instant domain augmentation for lidar semantic segmentation. In: Proceedings of the IEEE/CVF Conference on Computer Vision and Pattern Recognition. pp. 9350--9360 (2023)

\bibitem{saltori2022cosmix}
Saltori, C., Galasso, F., Fiameni, G., Sebe, N., Ricci, E., Poiesi, F.: Cosmix: Compositional semantic mix for domain adaptation in 3d lidar segmentation. In: European Conference on Computer Vision. pp. 586--602. Springer (2022)

\bibitem{shin_characteristics_2019}
Shin, J., Park, H., Kim, T.: Characteristics of laser backscattering intensity to detect frozen and wet surfaces on roads. Journal of Sensors  \textbf{2019}(1),  8973248 (2019)

\bibitem{smith_modeling_2018}
Smith, B.E., Gardner, A., Schneider, A., Flanner, M.: Modeling biases in laser-altimetry measurements caused by scattering of green light in snow. Remote sensing of environment  \textbf{215},  398--410 (2018)

\bibitem{thomas_kpconv_2019}
Thomas, H., Qi, C.R., Deschaud, J.E., Marcotegui, B., Goulette, F., Guibas, L.J.: Kpconv: Flexible and deformable convolution for point clouds. In: Proceedings of the IEEE/CVF international conference on computer vision. pp. 6411--6420 (2019)

\bibitem{tolstikhin_mlp-mixer_2021}
Tolstikhin, I.O., Houlsby, N., Kolesnikov, A., Beyer, L., Zhai, X., Unterthiner, T., Yung, J., Steiner, A., Keysers, D., Uszkoreit, J., et~al.: Mlp-mixer: An all-mlp architecture for vision. Advances in neural information processing systems  \textbf{34},  24261--24272 (2021)

\bibitem{xiao2022polarmix}
Xiao, A., Huang, J., Guan, D., Cui, K., Lu, S., Shao, L.: Polarmix: A general data augmentation technique for lidar point clouds. Advances in Neural Information Processing Systems  \textbf{35},  11035--11048 (2022)

\bibitem{xiao_polarmix_nodate}
Xiao, A., Huang, J., Guan, D., Cui, K., Lu, S., Shao, L.: Polarmix: A general data augmentation technique for lidar point clouds. Advances in Neural Information Processing Systems  \textbf{35},  11035--11048 (2022)

\bibitem{xiao_transfer_2022}
Xiao, A., Huang, J., Guan, D., Zhan, F., Lu, S.: Transfer learning from synthetic to real lidar point cloud for semantic segmentation. In: Proceedings of the AAAI conference on artificial intelligence. vol.~36, pp. 2795--2803 (2022)

\bibitem{xiao_3d_2023}
Xiao, A., Huang, J., Xuan, W., Ren, R., Liu, K., Guan, D., El~Saddik, A., Lu, S., Xing, E.P.: 3d semantic segmentation in the wild: Learning generalized models for adverse-condition point clouds. In: Proceedings of the IEEE/CVF Conference on Computer Vision and Pattern Recognition. pp. 9382--9392 (2023)

\bibitem{yan_benchmarking_2024}
Yan, X., Zheng, C., Xue, Y., Li, Z., Cui, S., Dai, D.: Benchmarking the robustness of lidar semantic segmentation models. International Journal of Computer Vision pp. 1--24 (2024)

\bibitem{yang_realistic_2023}
Yang, D., Liu, Z., Jiang, W., Yan, G., Gao, X., Shi, B., Liu, S., Cai, X.: Realistic rainy weather simulation for lidars in carla simulator. arXiv preprint arXiv:2312.12772  (2023)

\bibitem{zhao_point_2020}
Zhao, H., Jiang, L., Jia, J., Torr, P.H., Koltun, V.: Point transformer. In: Proceedings of the IEEE/CVF international conference on computer vision. pp. 16259--16268 (2021)

\bibitem{zhu_cylindrical_2021}
Zhu, X., Zhou, H., Wang, T., Hong, F., Ma, Y., Li, W., Li, H., Lin, D.: Cylindrical and asymmetrical 3d convolution networks for lidar segmentation. In: Proceedings of the IEEE/CVF conference on computer vision and pattern recognition. pp. 9939--9948 (2021)

\end{thebibliography}

\ifarXiv
    

\section*{Supplementary material}

# Rethinking Data Augmentation
# for Robust LiDAR Semantic Segmentation
# in Adverse Weather

## Supplementary Material

In this appendix, we supplement more materials to support the findings and conclusions in the main body of this paper. Specifically, this appendix is organized as follows. Section A provides additional explanations of Selective Jittering. Section B provides detailed explanations for Learnable Point Drop. Section C presents toy experiments conducted with SynLiDAR. Section D details class-level results. Section E offers further ablation studies. Section F provides additional qualitative results.

## A   Detailed Explanation for Selective Jittering

Selective Jittering is devised to address the primary distortion of geometric perturbation and consists of two types: Depth-Selective Jittering (DSJ) and Angle-Selective Jittering (ASJ). DSJ adds noise to points along discretized depth levels. The number of depth levels, randomly selected within a specific range, is a hyperparameter. ASJ perturbs points along angles, with the maximum angular range as a hyperparameter. In addition, Range Jittering (RJ) perturbs points in the range direction with Gaussian noise, which DSJ and ASJ do not.

## B   Detailed Explanation for Learnable Point Drop

(1) Concept: In Section 3, we claimed that adverse weather causes point drops and degrades performance. Therefore, once critical point drops that degrade the model performance are identified, we presume it can sufficiently mimic the effects of adverse weather. This is why LPD receives a reward as $L_{LPD} - L_{aug} + H_{LPD} - H_{aug}$, encouraged to find adverse point drops.

(2) Loss: The update loss in LPD is identical to that in the original DQN paper [5], with reward $r$, discount rate $\gamma$, frozen target model $Q'$ and policy model $Q$:

$$L_{DQN} = \mathbb{E}_{i,s,a} \left[ \left( r + \gamma \max_{a'} Q'(s', a'; \theta_{i-1}) - Q(s, a; \theta_i) \right)^2 \right]. \quad (1)$$

(3) The network needs only one backpropagation, and our training with LPD is performed only once. The loss of LPD from eqn. 1 and $L_{aug}$ and $L_{LPD}$ are combined and backpropagated together:

$$L_{total} = L_{aug} + L_{LPD} + L_{DQN}. \quad (2)$$

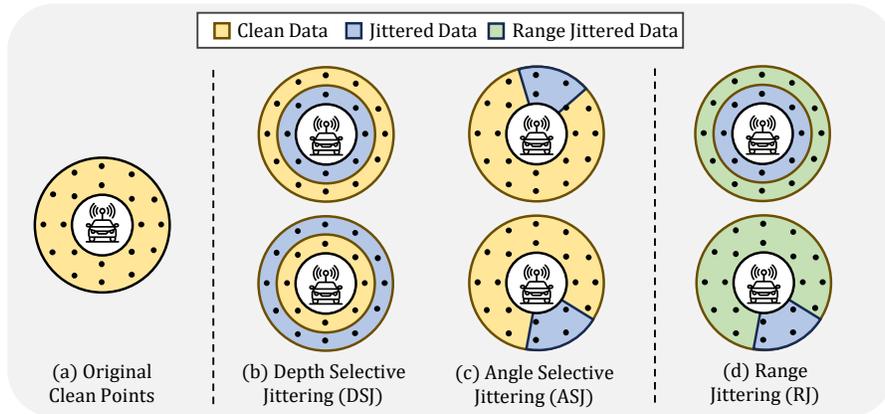

**Fig. A.1:** Illustration of proposed selective jittering. (a) Original points from clean data. (b) Depth-selective jittering (DSJ) adds Gaussian noise to a random range of depths. (c) Angle-selective jittering (ASJ) adds Gaussian noise to a random range of angles. (d) Range Jittering adds Gaussian noise in the range direction and is used alongside points treated with DSJ or ASJ.

## C  Toy Experiment in SynLiDAR

In the main paper, the toy experiment conducted on the SemanticKITTI [1] *validation set* was also performed using SynLiDAR [8]. As illustrated in Table A.1, the toy experiment in SynLiDAR yielded results akin to those observed in SemanticKITTI. Additionally, a detailed representation of the toy example's outcomes in SynLiDAR can be examined through Figure A.2.

**Table A.1:** Results of toy experiments from *validation set* of SynLIDAR [8].

| Methods | car | bi.cle | mt.cle | truck | oth-v. | pers. | bi.clst | mt.clst | road | parki. | sidew. | oth-g. | build. | fence | veget. | trunk | terra. | pole | traf. | mIoU |
|---|---|---|---|---|---|---|---|---|---|---|---|---|---|---|---|---|---|---|---|---|
| Clean | 0.0 | 54.9 | 94.1 | 0.0 | 0.0 | 94.8 | 93.7 | 0.0 | 94.8 | 0.0 | 89.8 | 0.0 | 98.2 | 53.7 | 76.8 | 90.3 | 0.0 | 86 | 95.2 | 53.8 |
| **D1** : soft | 0.0 | 37.0 | 86.7 | 0.0 | 0.0 | 89.5 | 87.8 | 0.0 | 92.6 | 0.0 | 85.1 | 0.0 | 93.8 | 32.4 | 39.5 | 80.9 | 0.0 | 84.7 | 93.0 | 50.2 |
| **D1** : hard | 0.0 | 9.1 | 53.9 | 0.0 | 0.0 | 56.0 | 48.6 | 0.0 | 20.9 | 0.0 | 23.7 | 0.0 | 31.0 | 0.4 | 4.8 | 26.7 | 0.0 | 39.6 | 74.4 | **21.6** |
| **D2** : soft | 0.0 | 4.2 | 86.7 | 0.0 | 0.0 | 88.2 | 87.8 | 0.0 | 92.5 | 0.0 | 48.4 | 0.0 | 52.1 | 7.0 | 1.4 | 68.8 | 0.0 | 28.8 | 26.8 | 31.2 |
| **D2** : hard | 0.0 | 0.2 | 52.5 | 0.0 | 0.0 | 50.3 | 48.4 | 0.0 | 20.6 | 0.0 | 3.1 | 0.0 | 4.0 | 0.2 | 0.1 | 10.5 | 0.0 | 2.2 | 8.7 | **11.1** |
| **D3** : soft | 0.0 | 55.7 | 79.6 | 0.0 | 0.0 | 86.5 | 87.4 | 0.0 | 45.0 | 0.0 | 50.4 | 0.0 | 55.3 | 22.3 | 6.8 | 74.1 | 0.0 | 69.5 | 29.1 | 34.8 |
| **D3** : hard | 0.0 | 12.5 | 11.3 | 0.0 | 0.0 | 63.3 | 32.9 | 0.0 | 0.2 | 0.0 | 3.5 | 0.0 | 6.6 | 1.9 | 2.8 | 45.6 | 0.0 | 4.1 | 0.1 | **9.7** |
| **D4** : soft | 0.0 | 54.8 | 94.1 | 0.0 | 0.0 | 94.8 | 93.8 | 0.0 | 94.8 | 0.0 | 89.8 | 0.0 | 98.2 | 53.8 | 76.6 | 90.2 | 0.0 | 86 | 95.2 | 53.8 |
| **D4** : hard | 0.0 | 54.8 | 94.1 | 0.0 | 0.0 | 94.8 | 93.8 | 0.0 | 94.8 | 0.0 | 89.8 | 0.0 | 98.2 | 53.8 | 76.6 | 90.2 | 0.0 | 86 | 95.2 | 53.8 |

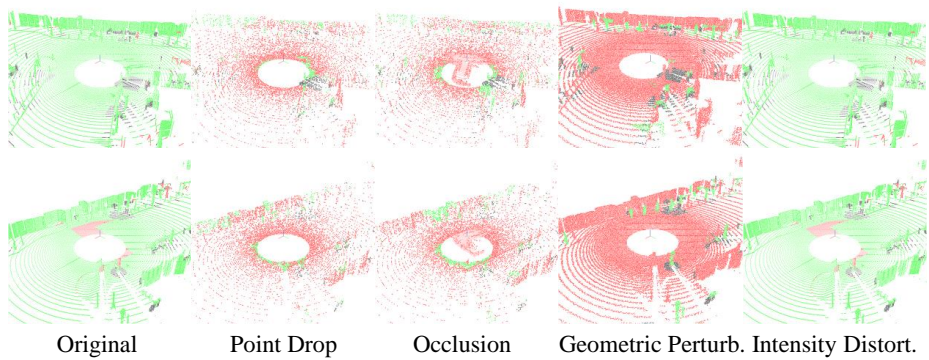

| | | | | | |
|---|---|---|---|---|---|
| Original | Point Drop | Occlusion | Geometric Perturb. | Intensity Distort. | |

**Fig. A.2:** Qualitative results of toy experiments in SynLiDAR [8]. MinkUnet is employed and trained in *train set* of the SynLiDAR dataset. The red points represent incorrect predictions, whereas the green points indicate correct predictions.

## D    Detailed Class-level Results

We provide detailed class IoU for the Table 2 of the main paper. As seen in Table A.2 for SemanticKITTI→SemanticSTF, our method has significantly improved the performance of categories such as *other vehicle*, *motorcyclist*, *sidewalk*, *pole*, and *traffic sign*, which initially had lower baseline performance. Specifically, it achieved increases of +12.0 IoU for *other vehicle*, +36.6 IoU for *motorcyclist*, +11.9 IoU for *sidewalk*, +4.9 IoU for *pole*, and +28.2 IoU for *traffic sign*. This demonstrates that our methodology is effective in predicting classes that are likely to be overlooked due to adverse weather conditions. Furthermore, the superiority of our method is evident as it shows better performance than the existing state-of-the-art method, PointDR, across most classes.

**Table A.2:** LiDAR segmentation results (mIoU) on the SemanticSTF validation set of models trained with SemanticKITTI. D-fog and L-fog denote dense fog and light fog weather conditions in all experiments. * is the reproduced result with the same segmentation backbone. The best score is in **bold** and the second best is <u>underlined</u>.

| Methods | car | bi.cle | mt.cle | truck | oth-v. | pers. | bi.clst | mt.clst | road | parki. | sidew. | oth-g. | build. | fence | veget. | trunk | terra. | pole | traf. | mIoU |
|---|---|---|---|---|---|---|---|---|---|---|---|---|---|---|---|---|---|---|---|---|
| Oracle | 89.4 | 42.1 | 0.0 | 59.9 | 61.2 | 69.6 | 39.0 | 0.0 | 82.2 | 21.5 | 58.2 | 45.6 | 86.1 | 63.6 | 80.2 | 52.0 | 77.6 | 50.1 | 61.7 | 54.7 |
| Baseline | 67.1 | <u>5.0</u> | **28.1** | 38.5 | <u>14.6</u> | 45.8 | 8.3 | 13.8 | 40.1 | **16.1** | 26.1 | 3.3 | 71.6 | **52.7** | 53.8 | 33.9 | 39.2 | <u>25.3</u> | 12.7 | 31.4 |
| LaserMix [4] | 18.6 | **5.4** | 0.0 | 9.9 | 1.6 | 0.6 | 7.9 | 10.5 | 47.6 | 6 | 12.1 | 1.8 | 21.6 | 20.2 | 48.4 | 6.6 | 37.8 | 19 | 2.8 | 14.7 |
| PolarMix [7] | 21 | 2 | 0.0 | 3.8 | 1.6 | 2.8 | 0.6 | 0.0 | <u>58.3</u> | 4.4 | 17.4 | 1.4 | 40.7 | 36.4 | 41.3 | 6.6 | 35 | 14.6 | 2.8 | 15.3 |
| PointDR* [9] | <u>69.2</u> | 1 | 8.9 | **41.9** | 7.6 | <u>48.9</u> | **17.0** | <u>36.2</u> | 57.8 | <u>15.9</u> | <u>32.3</u> | <u>4.0</u> | **75.7** | 46.4 | <u>54.0</u> | <u>36.2</u> | <u>43.9</u> | 23.7 | <u>24.2</u> | <u>33.9</u> |
| Baseline+SJ+LPD | **86.1** | 4.8 | <u>13.8</u> | <u>39.7</u> | **26.6** | **55.4** | <u>8.5</u> | **50.4** | **63.7** | 14.9 | **37.9** | **5.5** | <u>75.2</u> | **52.7** | **60.4** | **39.7** | **44.9** | **30.1** | **40.8** | **39.5** |
| Increments to baseline | +19.0 | -0.2 | -14.3 | +1.1 | +12.0 | +9.6 | +0.2 | +36.6 | +23.5 | -1.2 | +11.9 | +2.2 | +3.6 | 0.0 | +6.7 | +5.8 | +5.7 | +4.9 | +28.2 | +8.1 |

**Table A.3:** LiDAR segmentation results (mIoU) on the SemanticSTF validation set of models trained with SynLiDAR. D-fog and L-fog denote dense fog and light fog weather conditions in all experiments. * is the reproduced result with the same segmentation backbone. The symbol ‡ indicates that the validation for model selection was performed on sequence 0 of SynLiDAR, rather than SemanticSTF. The best score is in **bold** and the second best is <u>underlined</u>.

| Methods | car | bi.cle | mt.cle | truck | ot.h-v. | pers. | bi.clst | mt.clst | road | parki. | sidew. | ot.b-g. | build. | fence | veget. | trunk | terra. | pole | traf. | mIoU |
|---|---|---|---|---|---|---|---|---|---|---|---|---|---|---|---|---|---|---|---|---|
| Oracle | 89.4 | 42.1 | 0.0 | 59.9 | 61.2 | 69.6 | 39.0 | 0.0 | 82.2 | 21.5 | 58.2 | 45.6 | 86.1 | 63.6 | 80.2 | 52.0 | 77.6 | 50.1 | 61.7 | 54.7 |
| Baseline | 33.76 | 1.71 | <u>3.29</u> | 15.54 | 0.24 | 25.52 | 1.65 | 3.43 | 15.27 | 9.16 | 16.76 | 0.05 | 33.38 | 21.89 | 39.49 | 18.7 | **44.03** | 8.75 | 0.84 | 15.45 |
| LaserMix [4] | **52.41** | **5.6** | 1.05 | 7.91 | <u>1.96</u> | 25.59 | 1.62 | 2.92 | **44.58** | 6.44 | <u>21.21</u> | 0.88 | 16.95 | 23.61 | 40.75 | 18.91 | 41.45 | 5.65 | 0.7 | 16.85 |
| PolarMix [7] | <u>48.93</u> | <u>4.23</u> | 2.32 | 14.64 | **2.37** | 24.55 | 2.14 | 4.64 | 34.64 | 7.66 | 19.8 | 0.39 | 37.44 | 22.3 | 44.85 | 21.32 | <u>43.18</u> | 7.08 | 1.3 | 18.09 |
| PointDR* [7] | 41.07 | 2.81 | **3.43** | 18.05 | 0.17 | **31.3** | **2.81** | 3.3 | 34.39 | **10.23** | 19.72 | <u>0.96</u> | **52.72** | 21.98 | **48.49** | 21.33 | 38.31 | 19.19 | 5.61 | 19.78 |
| PointDR*‡ [9] | 36.13 | 3.47 | 2.15 | <u>21.93</u> | 0.31 | 28.72 | 1.69 | **5.09** | <u>42.92</u> | 9.31 | 20.71 | 0.58 | 50.83 | <u>26.88</u> | 46.85 | **24.49** | 37.69 | <u>22.74</u> | 6.45 | <u>20.47</u> |
| Baseline+SJ+LPD | 42.13 | 2.79 | 2.68 | 19.22 | 0.67 | <u>29.22</u> | 1.91 | <u>4.8</u> | 42.32 | 8.67 | 21.05 | **1.56** | 48.23 | 25.97 | 47.17 | 22.11 | 32.8 | 21.74 | <u>6.54</u> | 20.08 |
| Increments to baseline | +8.4 | +1.1 | -0.6 | +3.7 | +0.4 | +3.7 | +0.3 | +1.4 | +27.0 | -0.5 | +4.3 | +1.5 | +14.9 | +4.1 | +7.7 | +3.4 | -11.2 | +13.0 | +5.7 | +4.6 |
| Baseline+SJ+LPD‡ | 39.26 | 2.89 | 0.89 | **19.39** | 0.75 | 27.68 | **2.19** | 3.78 | 42.5 | <u>9.35</u> | **21.55** | 0.3 | <u>51.89</u> | **33.48** | <u>47.38</u> | <u>23.11</u> | 33.31 | **23.22** | **6.78** | **20.51** |
| Increments to baseline | +5.5 | +1.2 | -2.4 | +3.8 | +0.5 | +2.2 | +0.5 | +0.4 | +27.2 | +0.2 | +4.8 | +0.3 | +18.5 | +11.6 | +7.9 | +4.4 | -10.7 | +14.5 | +5.9 | +5.1 |

In the experiment for SynLiDAR→SemanticSTF, our methodology significantly improves performance on key classes in a driving environment, increasing the IoU for 'car' by +5.5, 'person' by +2.2, and 'road' by +27.2 as shown in Tab. A.3. Additionally, it shows substantial performance gains compared to the baseline, with increases of +4.8 IoU for 'sidewalk', +18.5 IoU for 'building', and +11.6 IoU for 'fence'. This indicates that our methodology effectively aids in predicting classes that become challenging to forecast due to adverse weather conditions.

# E   More Ablation Studies

**Analysis of DSJ and ASJ.** Our study compared DSJ and ASJ against jittering applied to all points. Additionally, to evaluate Cube-Selective Jittering (CSJ), which involves jittering points within a random cubic area. According to Tab. A.4, jittering all points improved robustness in +5.5 mIoU but resulted in a -2.2 mIoU performance decrease on clean data. In contrast, CSJ led to a decrease in mIoU across all adverse weather conditions in a -0.2 mIoU. This outcome suggests that not considering the prior spherical shape of LiDAR data when choosing jittering locations may be inadequate for countering adverse weather. Conversely, ASJ demonstrated improvements (+2.6 mIoU in dense fog, +5.3 mIoU in light fog, +7.9 mIoU in rain, and +6.3 mIoU in snow) with a relatively minor -1.8 mIoU performance decrease in clean data. Also, DSJ showed similar trends with increases in mIoU across various adverse weather conditions and a more moderate -1.7 mIoU drop in clean data performance. These results indicate that ASJ and DSJ effectively simulate coordinate distortion while minimizing the performance gap with clean data.

**Table A.4:** Experiments on ASJ/DSJ relative to all-point jittering/CSJ. All models were trained on SemanticKITTI and validated on SemanticSTF. The values in parentheses indicate the performance improvement or decrease over the baseline model.

| Methods | Clean | D-fog | L-fog | Rain | Snow | mIoU |
|---|---|---|---|---|---|---|
| Baseline | **63.9** | 30.7 | 30.1 | 29.7 | 25.3 | 31.4 |
| Jittering | 61.7 (-2.2) | <u>34.3</u> | **36.3** | 33.2 | **32.4** | <u>36.9</u> (+5.5) |
| CSJ | <u>63.2</u> (-0.7) | 32.9 | 31.1 | 28.8 | 23.2 | 31.2 (-0.2) |
| ASJ | 62.1 (-1.8) | 33.3 | 35.4 | **37.8** | <u>31.6</u> | 36.8 (+5.4) |
| DSJ | 62.2 (-1.7) | **34.6** | **36.8** | <u>36.1</u> | 31.2 | **38.0** (+6.6) |

**Analysis of LPD.** In our research, we investigated how LPD differs from random dropout. As indicated in Tab. A.5, LPD increasingly drops points as the distance increases. Fog tends to obscure objects that are further away more significantly. Therefore, this demonstrates that LPD effectively represents point drop due to occlusions such as fog, which it is designed to simulate. Consequently, LPD is a more reasonable point dropout method to deal with adverse weather conditions than random point drop.

**Table A.5:** Ratio of points remaining after LPD compared to original data. This contrasts with random drop, which exhibits a uniform remaining ratio across different distances.

| Distance | 0~10m | 10~20m | 20~30m | 30~40m | 40~50m | 50~60m | 60~70m | 70~80m | 80~90m |
|---|---|---|---|---|---|---|---|---|---|
| Ratio | 80.6 | 78.6 | 78.7 | 77.4 | 76.2 | 75.5 | 76.2 | 76.4 | 72.5 |

**Augmentation Configurations and Parameter Ablation.** (1) DSJ adds noise to points along discretized depth levels. The number of depth levels, randomly selected within a specific range, is a hyperparameter. More depth levels improved rain resistance. (2) ASJ perturbs points along angles, with the maximum angular range $\Delta\theta$ as a hyperparameter. A larger angle range improved dense fog resistance. (3) Gaussian noise determines the jittering severity in SJ with its standard deviation $\sigma$ as a hyperparameter. Higher $\sigma$ improved resistance to rain and snow; lower $\sigma$ to fog. (4) LPD involves a batch size, discount rate $\gamma$, and exploration decay $\epsilon$ as hyperparameters related to DQN learning. Higher $\gamma$ improved fog resistance. Ablation studies in Tab. A.6 show that we achieve generalized performance, less sensitive to hyperparameter choices.

**Comparison with Domain Generalization and Adaptation Methods.** CoSMix [6] and UniMix [10] offer unsupervised domain adaptation (DA) techniques, with UniMix also providing domain generalization (DG) techniques. Domain generalization methods train models using only source domain data, while

**Table A.6:** Hyperparameter ablation in SemanticKITTI-to-SemanticSTF. The values we chose are highlighted in **bold**.

| | Selective Jittering | | | | | | Learnable Point Drop | | | | |
|---|---|---|---|---|---|---|---|---|---|---|---|
| DSJ range | mIoU | ASJ $\Delta\theta$ | mIoU | Gaussian $\sigma$ | mIoU | Batch size | mIoU | Discount $\gamma$ | mIoU | Decay $\epsilon$ | mIoU |
| [2, 5] | 39.3 | $\frac{1}{16}\pi$ | 38.1 | 0.001 | 37.5 | 8 | 37.1 | 0.5 | 37.8 | 100 | 38.5 |
| [3, 6] | 38.8 | $\frac{1}{8}\pi$ | 39.9 | 0.005 | 38.4 | 16 | 37.1 | 0.8 | 38.4 | 500 | 38.7 |
| [4, 7] | 37.8 | $\pi$ | **39.5** | **0.01** | **39.5** | **32** | **39.5** | 0.9 | 38.4 | **1000** | **39.5** |
| **[5,8]** | **39.5** | $\frac{1}{2}\pi$ | 40.0 | 0.05 | 37.3 | 64 | 38.2 | **0.99** | **39.5** | 2000 | 37.3 |
| [6,9] | 37.8 | $\frac{3}{4}\pi$ | 39.3 | 0.1 | 37.9 | 128 | 37.0 | 0.999 | 38.8 | 3000 | 40.3 |
| deviation | 0.8 | deviation | 0.2 | deviation | 1.3 | deviation | 1.7 | deviation | 0.9 | deviation | 0.7 |

unsupervised domain adaptation methods train models with unlabeled target domain data. Our method, however, exclusively uses source data, making it fundamentally more challenging than UDA and not directly comparable. As shown in Tab. A.7, our method outperforms CosMix and UniMix (DG) by +7.9 and +4.9 mIoU, respectively. This result demonstrates that our proposed method effectively represents adverse weather conditions without real weather-conditioned data. Additionally, our method performs comparably to UniMix (DA) despite not using target domain data as DA does. Because UDA can complement data augmentations, combining our method with UniMix could further enhance performance.

**Table A.7:** Comparison with CosMix and UniMix. DA and DG denote domain adaptation and domain generalization.

| Methods | mIoU |
|---|---|
| CoSMix(DA) | 28.4 |
| UniMix(DA) | 37.2 |
| UniMix(DG) | 31.4 |
| MinkowskiNet18+Ours | 36.3 |

**Time cost of augmentation process.** First, we emphasize that the inference runtime does not change. Regarding the training overhead, it takes 3.4 hours to train the original MinkowskiNet 15 epochs with four A6000 GPUs, and our augmentations take an extra 2.28 hours (+ 67%). In comparison, existing weather simulations [2,3] took over 12 hours for SemanticKITTI's train set, highlighting our method's efficiency relative to traditional simulations.

# F    More Qualitative Results

The figures below show the qualitative examples of LiDAR semantic segmentation results on the SemanticSTF validation set.

As observed in Fig. A.3, proposed SJ and LPD were found to improve the prediction of roads and sidewalks in dense fog conditions. Similarly, in light fog conditions, as demonstrated in the third and fourth rows of Fig. A.3, there was a significant enhancement in the performance for road predictions. Moreover, as shown in the black circle in the third row of Fig. A.3, accurate predictions were also achieved for road-related classes such as cars. This trend of improved performance under adverse weather conditions continued, as can be seen in Fig. A.4.

This observed trend of improved performance persists even when training on SynLiDAR and performing validation on SemanticSTF. As seen in Fig. A.5, our method successfully executes accurate predictions in locations where the baseline model previously erred. Similarly, in the third row of Fig. A.5, enhanced precision in predicting objects like cars is evident. This consistent trend of enhanced accuracy under various weather conditions can also be found in Fig. A.6.

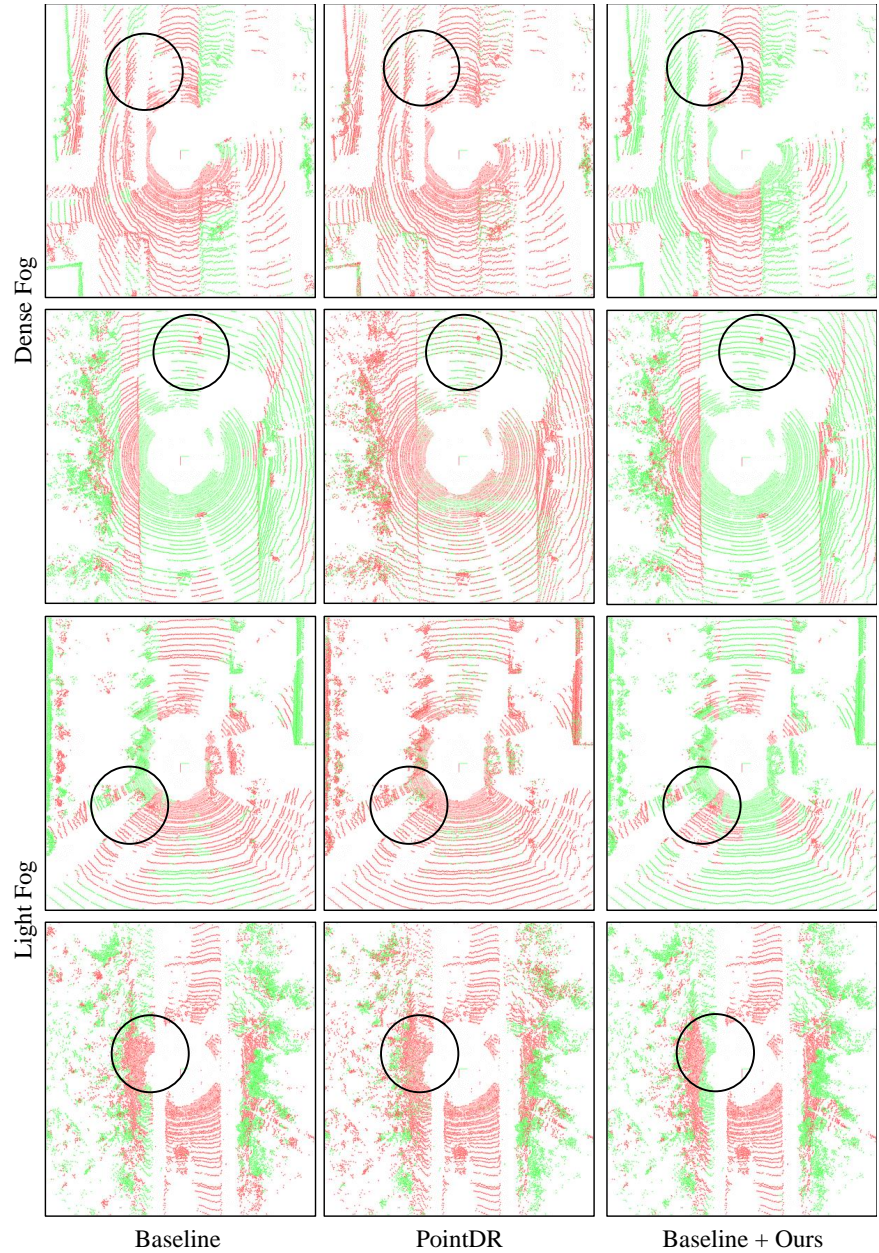

**Fig. A.3:** Qualitative examples of LiDAR semantic segmentation result on the **_dense fog_** and **_light fog_** condition of the SemanticSTF _validation set._ All models are trained on SemanticKITTI. In all qualitative results, the red and green points indicate incorrect and correct predictions respectively. Best viewed in colors.

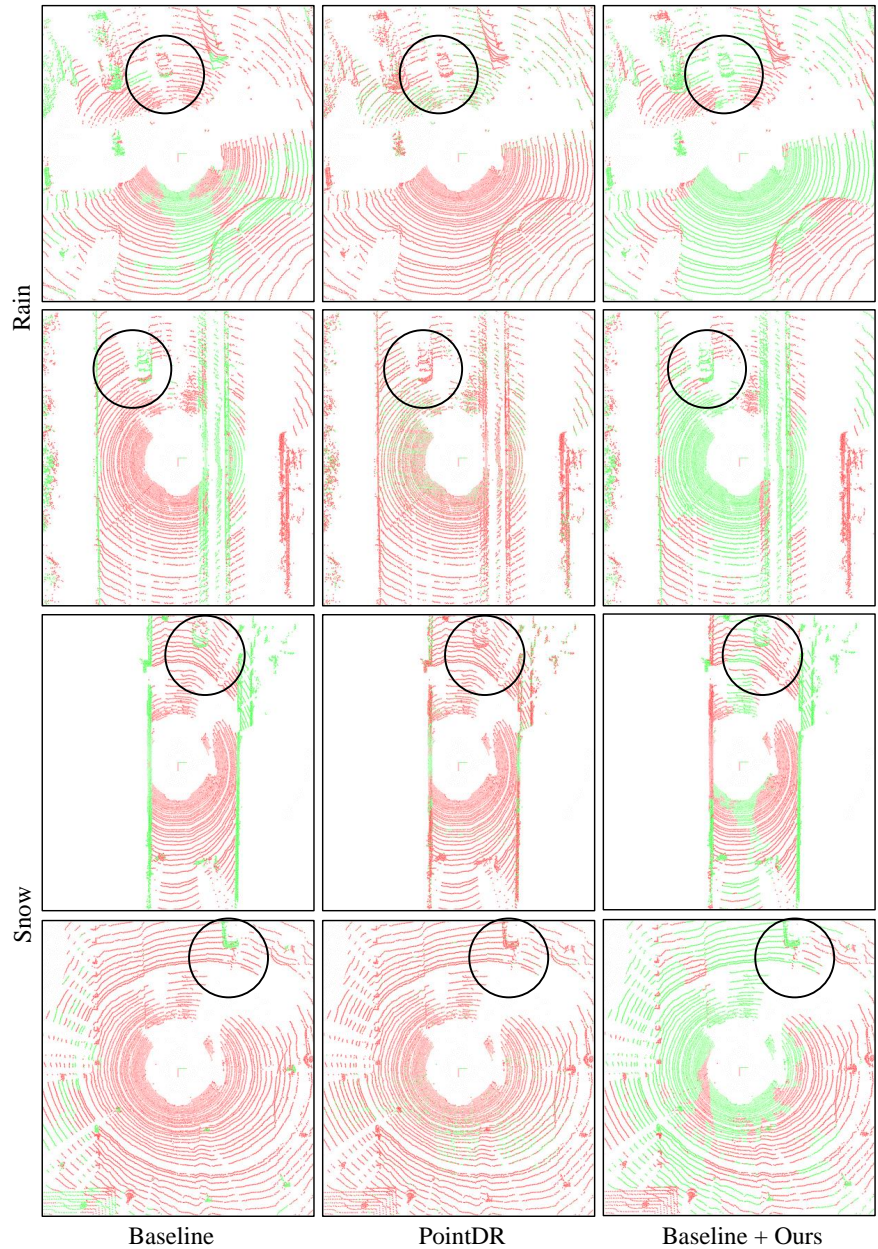

**Fig. A.4:** Qualitative examples of LiDAR semantic segmentation result on the **rain** and **snow** condition of the SemanticSTF *validation set*. All models are trained on SemanticKITTI. In all qualitative results, the red and green points indicate incorrect and correct predictions respectively. Best viewed in colors.

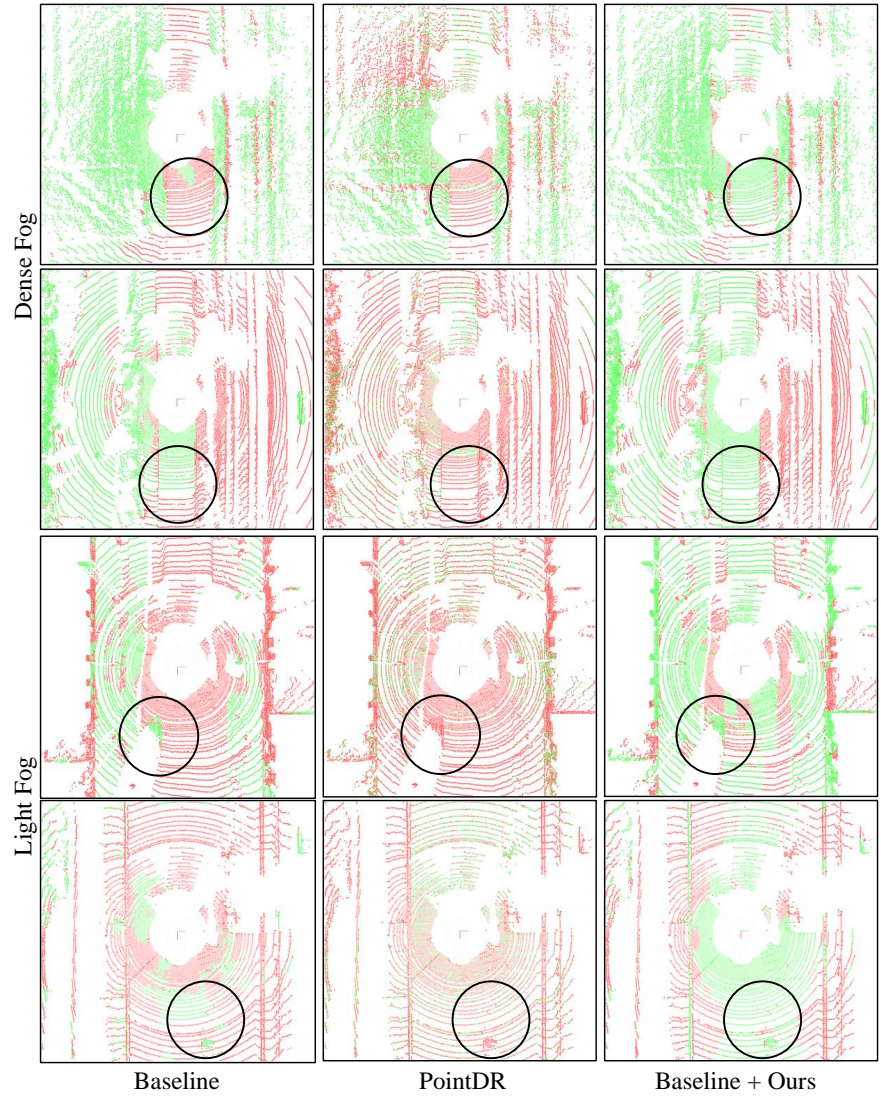

**Fig. A.5:** Qualitative examples of LiDAR semantic segmentation result on the ***dense fog*** and ***light fog*** condition of the SemanticSTF *validation set*. All models are trained on SynLiDAR. In all qualitative results, the red and green points indicate incorrect and correct predictions respectively. Best viewed in colors.

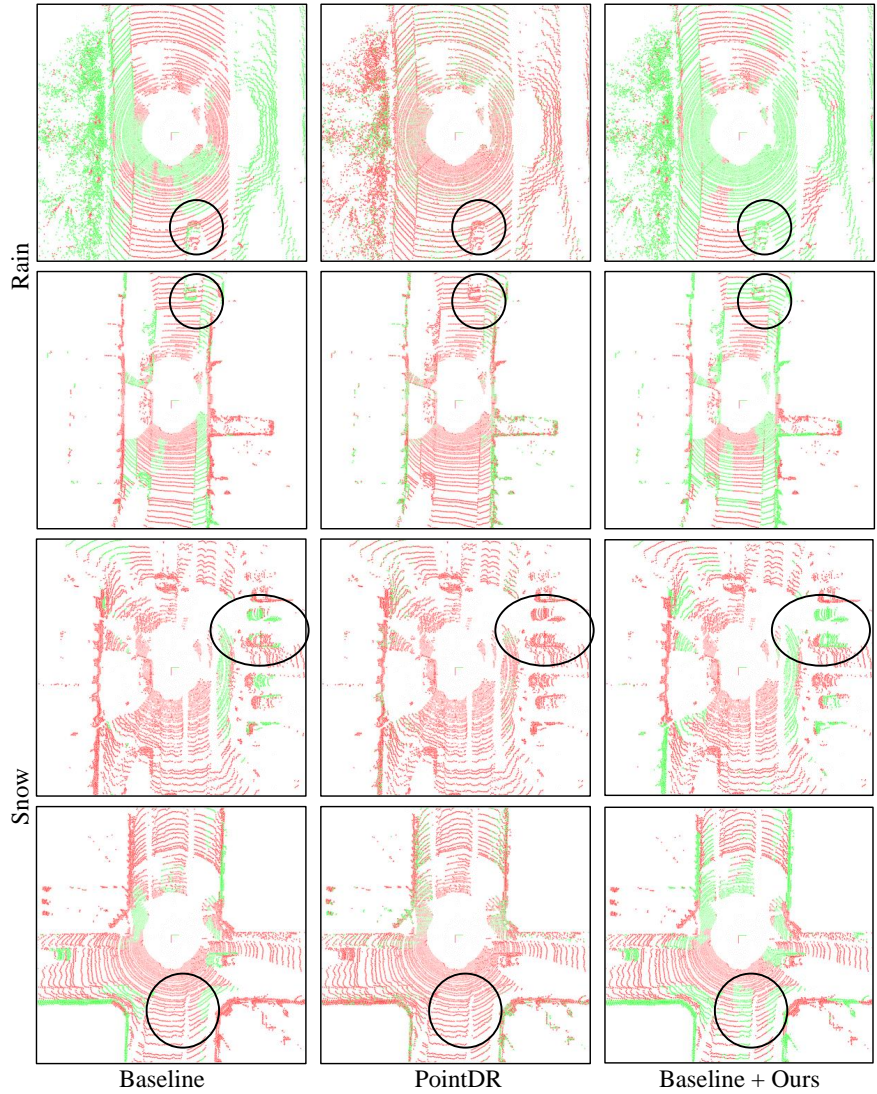

**Fig. A.6:** Qualitative examples of LiDAR semantic segmentation result on the ***rain*** and ***snow*** condition of the SemanticSTF *validation set.*. All models are trained on SynLiDAR. In all qualitative results, the red and green points indicate incorrect and correct predictions respectively. Best viewed in colors.


\fi

\end{document}